\documentclass[letterpaper]{article} %
\usepackage[preprint]{aaai2027}  %
\usepackage[hyphens]{url}  %
\usepackage{graphicx} %
\usepackage{natbib}  %
\usepackage{caption} %
\usepackage{booktabs}
\usepackage{subcaption}
\usepackage{arydshln}
\usepackage{multirow}
\usepackage{siunitx}
\usepackage{makecell}
\usepackage{bm}
\usepackage{amsmath}
\usepackage{amssymb}
\usepackage{dsfont}   %
\usepackage{mathtools}
\usepackage{amsthm}
\usepackage{gensymb}
\usepackage{enumitem}
\usepackage{algorithm}
\usepackage{algorithmic}
\usepackage{xcolor}
\usepackage{colortbl}
\usepackage{xspace}

\makeatletter
\DeclareRobustCommand\onedot{\futurelet\@let@token\@onedot}
\def\@onedot{\ifx\@let@token.\else.\null\fi\xspace}

\makeatother

\newcommand{\beginsupplement}{%
\setcounter{table}{0}
\renewcommand{\thetable}{S\arabic{table}}%
\setcounter{figure}{0}
\renewcommand{\thefigure}{S\arabic{figure}}%
}

\newcommand{\xclipgs}{XClipGS\xspace}
\newcommand{\dgs}{N-DGS\xspace}

\newcommand{\opCull}{HC\xspace}       %
\newcommand{\opMoment}{MM\xspace}     %
\newcommand{\opAnalytic}{Ours\xspace} %

\newcommand{\ClipGSgelpsnr}{32.20}\newcommand{\ClipGSgelssim}{.960}\newcommand{\ClipGSgellpips}{.089}
\newcommand{\ClipGSintestinepsnr}{30.83}\newcommand{\ClipGSintestinessim}{.944}\newcommand{\ClipGSintestinelpips}{.100}
\newcommand{\ClipGSkneejointpsnr}{27.89}\newcommand{\ClipGSkneejointssim}{.904}\newcommand{\ClipGSkneejointlpips}{.197}
\newcommand{\ClipGSlowerpsnr}{33.02}\newcommand{\ClipGSlowerssim}{.977}\newcommand{\ClipGSlowerlpips}{.028}
\newcommand{\ClipGSvascularpsnr}{34.68}\newcommand{\ClipGSvascularssim}{.971}\newcommand{\ClipGSvascularlpips}{.035}
\newcommand{\ClipGSheartpsnr}{28.14}\newcommand{\ClipGSheartssim}{.930}\newcommand{\ClipGSheartlpips}{.106}
\newcommand{\ClipGSnosepsnr}{34.65}\newcommand{\ClipGSnosessim}{.973}\newcommand{\ClipGSnoselpips}{.051}
\newcommand{\ClipGShandpsnr}{37.30}\newcommand{\ClipGShandssim}{.974}\newcommand{\ClipGShandlpips}{.041}
\newcommand{\ClipGSgelfps}{293}\newcommand{\ClipGSintestinefps}{220}\newcommand{\ClipGSkneejointfps}{252}
\newcommand{\ClipGSlowerfps}{241}\newcommand{\ClipGSvascularfps}{320}\newcommand{\ClipGSheartfps}{239}
\newcommand{\ClipGSnosefps}{306}\newcommand{\ClipGShandfps}{353}
\newcommand{\ClipGSpsnr}{32.34}\newcommand{\ClipGSssim}{.954}\newcommand{\ClipGSlpips}{.081}\newcommand{\ClipGSfps}{278}
\newcommand{\ClipGSgap}{+1.22~dB}  %

\newcommand{\bandS}{S$_{\mathrm{b}}$}      %
\newcommand{\cdem}{CDE}                    %
\newcommand{\cdemG}{CDE\textsubscript{g}}  %
\newcommand{\cdemP}{CDE\textsubscript{p}}  %
\newcommand{\leakm}{Leak}                  %
\newcommand{\cerrD}{CErr\textsubscript{3D}}

\newcommand{\mcdem}{\mathrm{CDE}}
\newcommand{\mleak}{\mathrm{Leak}}
\newcommand{\mcerrD}{\mathrm{CErr_{3D}}}
\newcommand{\rwparagraph}[1]{\vspace{-3pt}\paragraph{#1}}

\theoremstyle{plain}
\newtheorem{theorem}{Theorem}
\newtheorem{proposition}[theorem]{Proposition}

\theoremstyle{definition}

\theoremstyle{remark}

\title{\xclipgs{}: Exact Half-Space Clipping for Medical Volume Gaussian Splatting}
\author{
    Zhongpai Gao\textsuperscript{\rm 1,*},
    Benjamin Planche\textsuperscript{\rm 1},
    Meng Zheng\textsuperscript{\rm 1},
    Anwesa Choudhuri\textsuperscript{\rm 1},
    Chaoyi Zhou\textsuperscript{\rm 1,2,\dag},\\
    Terrence Chen\textsuperscript{\rm 1},
    Ziyan Wu\textsuperscript{\rm 1}
}
\affiliations{
    \textsuperscript{\rm 1}United Imaging Intelligence, Boston MA, USA\quad
    \textsuperscript{\rm 2}Clemson University, Clemson SC, USA\\
    zhongpai.gao@uii-ai.com
}

\newcommand{\xclipgsteaser}{%
\begin{minipage}{\textwidth}
\centering
\includegraphics[width=\textwidth]{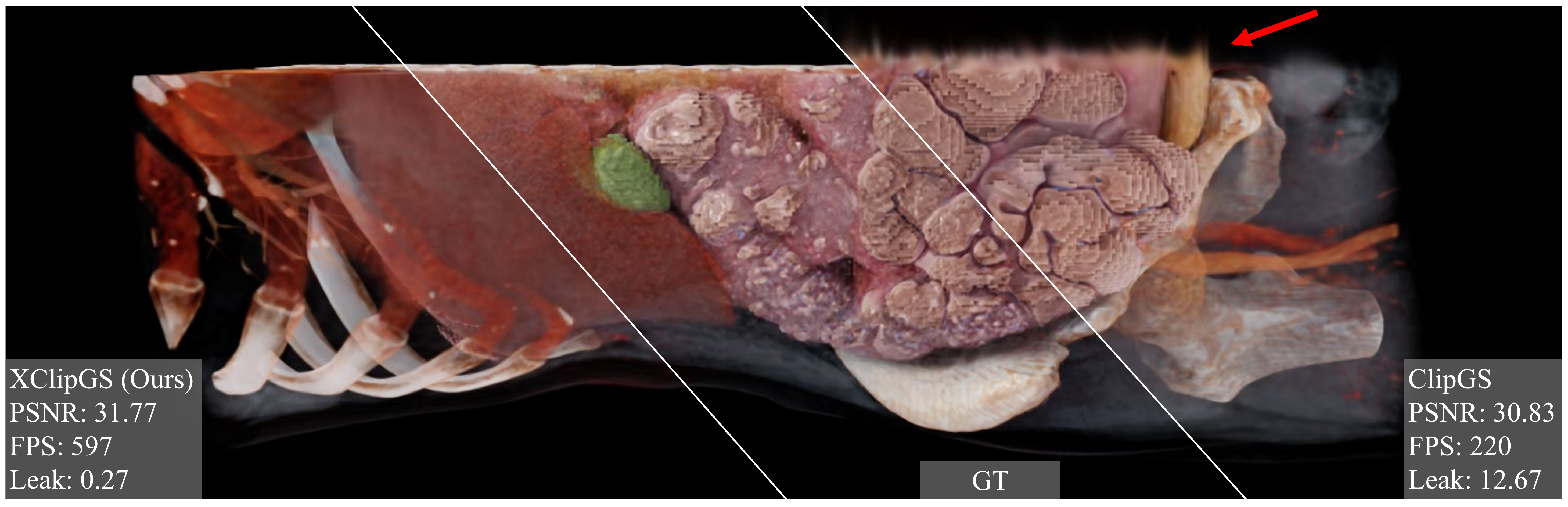}
\captionof{figure}{\textbf{Faithful interactive clipping.}
XClipGS, reference volume rendering, and ClipGS~\citep{li2025clipgs} (left to right) on a held-out test
cut. The arrow marks ClipGS leakage; labels report quality and speed.}
\label{fig:teaser}
\end{minipage}%
}

\makeatletter
\let\xclipgs@originalmaketitle\@maketitle
\def\@maketitle{%
  \xclipgs@originalmaketitle
  \vspace{-0.75em}%
  \xclipgsteaser
  \vspace{0.65em}%
}
\makeatother

\begin{document}

\raggedbottom
\maketitle

\begingroup
\renewcommand\thefootnote{}
\footnotetext{\textsuperscript{*}Corresponding author.\quad
\textsuperscript{\dag}The contribution of this author was made while interning at United Imaging Intelligence.}
\addtocounter{footnote}{-1}
\endgroup

\begin{abstract}
Gaussian-splatting proxies enable interactive rendering of volumetric medical
scans, but a clipping plane exposes anatomy not constrained by external-view
training and intersects primitives that conventional splatting can only keep or
drop whole. We present \xclipgs{} (\emph{eXact Clipping}), which treats these as
two separate problems: the render-time clip operator and supervision of the
hidden interior. Under the local affine model used by EWA splatting, each
half-space-restricted Gaussian's ray integral factorizes exactly into its ordinary
2D footprint and a conditional Gaussian CDF whose argument is affine in pixel
coordinates. The resulting closed-form per-splat operator introduces no learned clipping parameters or
auxiliary network and remains differentiable with respect to the primitive and
plane. We use multi-distance reference views with varied clipping-plane axes and
offsets to supervise the interior through the same operator. We also introduce a
paired clipped/unclipped cut-face protocol with difference-referenced cut error
(\cdem{}) and culled-side leakage (\leakm{}), because global image metrics dilute
errors near the plane. On eight CT and MRI volumes with plane offsets not used for training,
\xclipgs{} attains the highest PSNR on every volume ($33.56$ versus $32.34$~dB for
ClipGS) while rendering at over $650$~FPS, far above real
time, versus $278$~FPS. On voxel-axis cut-face views it raises average band
SSIM from $.809$ to $.860$ and leaks roughly $41\times$ less. Without
retraining, it also achieves the best average across all four metrics on
arbitrary-normal planes; on a fixed interior it matches RaRa's face fidelity
with about $16\times$ less leakage.
Project page: \url{https://gaozhongpai.github.io/XClipGS/}.
\end{abstract}

\section{Introduction}
\label{sec:introduction}

CT and MRI are inspected by orbiting, zooming, and clipping the volume to expose
hidden anatomy~\citep{weiskopf2003clipping,niedermayr2024application}. Cinematic volume rendering makes
these cross-sections legible through realistic depth and material cues, but
converged rendering remains expensive on commodity and web hardware. Gaussian
splatting~\citep{3dgs} offers a real-time proxy: a volume can be
represented by compact primitives and rasterized at interactive rates. Existing
medical proxies are usually trained from views surrounding the intact volume,
however, so their supervision concentrates on the visible outer anatomy.

An interactive clipping plane changes the rendering problem. It removes the
outer layers and turns a previously hidden cross-section into the dominant
surface. A proxy that matches external views can therefore fail abruptly when a
user moves the plane through its interior. This challenges representation
generalization and requires the renderer to determine each intersected
Gaussian's partial contribution.

We separate these effects into two failure modes. The \textbf{clip operator} must truncate
primitives intersected by the plane: binary keep/drop schemes, including
ClipGS's trainable decision~\citep{li2025clipgs}, quantize the boundary, while
RaRa's ray--ellipsoid chord ratio~\citep{li2025rara} only approximates partial
contribution. The
\textbf{unsupervised interior} is equally important: an accurate operator
reveals whatever anatomy the proxy learned behind the outer surface. External
views can leave that anatomy ambiguous, while supervision through an
approximate operator can teach the proxy the operator's boundary errors.
Faithful clipping therefore requires both an accurate partial-primitive
rule and training views that reveal the hidden volume: with the exact operator in
place, removing the clipped training views still costs $8.3$~dB on held-out
clipped views across all eight volumes.

\xclipgs{} (\emph{eXact Clipping}) addresses both with a closed-form,
differentiable per-pixel clip and clip-aware interior supervision
(Figure~\ref{fig:teaser}). Under the local affine model of EWA
splatting~\citep{zwicker2001ewa}, each half-space-restricted Gaussian's ray
integral equals its ordinary 2D footprint times a conditional Gaussian CDF
whose argument is affine in pixel position. The clip is exact under affine EWA
at the per-splat integration level and requires no learned clipping parameters
or auxiliary network. Rendering the
proxy through this operator against clipped reference views supplies direct
gradients to primitives near the exposed face. We vary camera distance, plane axis, and offset so
that training includes full-volume context and detailed, diverse
cross-sections. The construction acts on primitive geometry and applies
unchanged to CT and MRI within the same training and rendering pipeline.

Our experiments distinguish whole-system quality from operator quality. We
compare the analytic operator with a hard cull (HC) and the forward-KL-optimal
moment-matched Gaussian surrogate (MM) under a shared
\dgs{} backbone~\citep{gao20246dgs}. A
3DGS-based ClipGS~\citep{li2025clipgs} provides a system comparison, while operator
swaps on fixed checkpoints isolate the render-time rule and include RaRa~\citep{li2025rara}.
Numerically held-out plane offsets test interpolation along the sampled voxel
axes, while a separate test uses arbitrary plane orientations without
retraining. Paired clipped and unclipped
reference renders localize error at the cut face, where global metrics
are least informative.

In summary, this paper contributes:
\begin{itemize}[leftmargin=1.4em,itemsep=1pt,topsep=2pt]
\item \textbf{Analytic half-space clipping}: a closed-form, differentiable
per-pixel Gaussian integral that is exact for each primitive under the affine
EWA model and
introduces no learned clipping parameters or auxiliary network;
\item \textbf{Clip-aware interior supervision}: multi-distance reference views
whose clipping-plane axes and offsets vary across the volume, with evaluation on
held-out offsets and arbitrary orientations absent from training, testing transfer
to new cross-sections without retraining;
\item \textbf{A CT and MRI benchmark with cut-face metrics}: eight
volumes with held-out planes and metrics that score the cut where global
image quality is nearly blind to it, on which our method achieves the best
average on all four metrics for both plane sets. Parameter sweeps verify that
localization and leakage conclusions are stable across thresholds, band widths,
and leakage-mask settings.
\end{itemize}

\section{Related Work}
\label{sec:related}

\rwparagraph{Medical volume rendering and Gaussian proxies.}
Cinematic medical volume rendering path-traces global illumination through the
volume~\citep{kroes2012exposure,comaniciu2016shaping,novak2018monte}, making
cross-sections legible through realistic depth and material
cues~\citep{dappa2016cinematic}. Converged rendering is not interactive,
progressive estimates can flicker~\citep{martschinke2019adaptive}, and web
deployment remains difficult~\citep{xu2022inbrowser}.
\citet{niedermayr2024application} reproduce cinematic anatomy in real time with
a Gaussian proxy, but its clip planes are fixed during preprocessing. Other
Gaussian methods target rendering, reconstruction, or novel-view
synthesis~\citep{gao2024ddgs,cai2024xgaussian,zha2024r2gaussian,yang2024deform3dgs,peng2025mrigaussian,yang2026exactgs},
not partial-primitive clipping.

\rwparagraph{Geometric clipping of Gaussian splats.}
ClipGS~\citep{li2025clipgs}, our primary medical baseline, makes a binary
per-primitive decision trainable through an STE and deforms primitives with a
plane-conditioned MLP. RaRa Clipper~\citep{li2025rara} scales intersecting
primitives by the visible ray--ellipsoid chord fraction, and its released
formulation has no backward pass for clip-aware training. Earlier medical work
toggles opacity layers~\citep{kleinbeck2024multilayer}. Our operator instead
integrates the half-space-restricted Gaussian ray density under affine EWA and
supports both supervision and rendering. Section~\ref{sec:experiments} compares
the system with ClipGS and
isolates RaRa as a render-time rule on a fixed interior.

\rwparagraph{\mbox{Conditioned Gaussian splatting.}}\hspace{-0.5em}
6DGS~\citep{gao20246dgs} and 7DGS~\citep{gao20257dgs} form N-dimensional
Gaussian splatting (N-DGS), with
Beta-conditioned generalizations~\citep{liu2025universal,liu2025deformable}.
Our opacity-conditioned \dgs{} backbone leaves each primitive an ordinary 3D
Gaussian, so geometric clipping applies unchanged; its statistical ``slicing''
is distinct from a planar cut. Render-FM~\citep{gao2025renderfm} supplies
initialization.

\rwparagraph{Gaussian-splatting fidelity.}
Our derivation adopts the local affine EWA projection used by
3DGS~\citep{3dgs,zwicker2001ewa}. Exact perspective-ray methods such as
3DGEER~\citep{huang2026threedgeer} change the projection model and require
re-derived coefficients. Our exactness claim accordingly concerns the
per-primitive affine-EWA ray integral; the base renderer's splat order and alpha
compositing remain unchanged. Analytic-Splatting integrates over projected
pixels, while Mip-NeRF and Mip-Splatting address pixel-footprint
sampling~\citep{liang2024analyticsplatting,barron2021mipnerf,yu2023mipsplattingaliasfree3dgaussian};
these concerns are complementary to half-space restriction.

\rwparagraph{Semantic editing and half-space representations.}
Semantic editors select or remove whole primitive
subsets~\citep{chen2024gaussianeditor,ye2024gaussiangrouping,cen2025saga,zhou2024feature3dgs,jain2024gaussiancut}.
3D-HGS instead learns a fixed half-Gaussian split~\citep{li2024halfgaussian};
neither evaluates the continuous contribution of a user-supplied plane.
EVSplitting replaces both sides with moment-preserving
Gaussians~\citep{feng2024evsplitting}. Like our truncated-normal MM
ablation~\citep{tallis1961moment}, it re-Gaussianizes; our operator
integrates the restricted density.

\section{Method}
\label{sec:method}

\begin{figure*}[t]
\centering
\includegraphics[width=\textwidth]{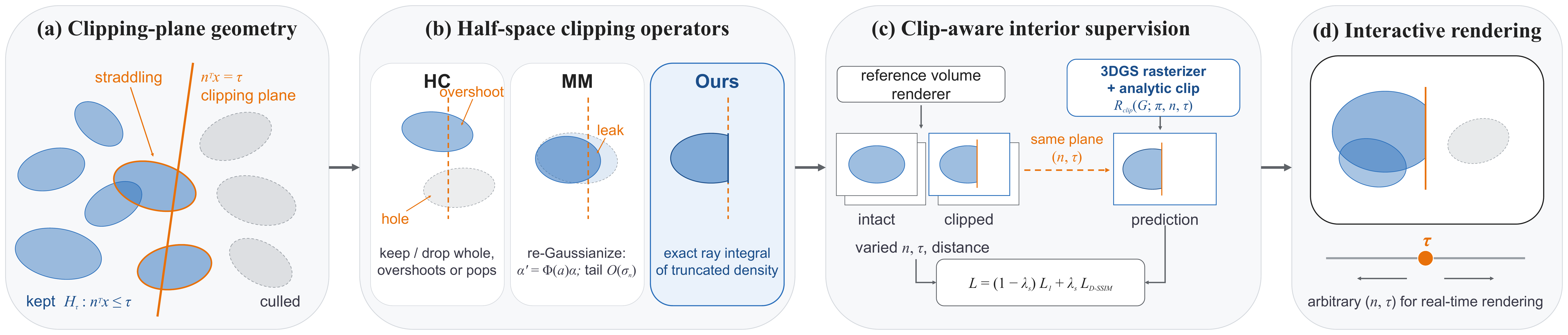}
\caption{\textbf{XClipGS pipeline.}
Exact per-splat half-space integration clips straddling Gaussians, while multi-distance
cut views supervise the hidden interior through the same differentiable operator.}
\label{fig:method-overview}
\vspace{-1em}
\end{figure*}

\paragraph{Overview.}
XClipGS combines an exact per-splat render-time clipping operator with clip-aware
supervision that learns the hidden interior (Figure~\ref{fig:method-overview}).
The operator analytically integrates plane-straddling Gaussians under affine
EWA, while multi-distance clipped views with varied plane axes and offsets train
the interior through the same differentiable rule. We first establish the
rendering model, derive the analytic operator and its hard-cull and
moment-matched reference points, and then describe interior supervision.

\subsection{Setup and notation}
\label{subsec:setup}

A scene is a set of Gaussian primitives $\{(\boldsymbol{\mu}_i, \boldsymbol{\Sigma}_i, \alpha_i, \mathbf{c}_i)\}_{i=1}^{N}$ with mean $\boldsymbol{\mu}_i\in\mathbb{R}^3$, covariance $\boldsymbol{\Sigma}_i\in\mathbb{R}^{3\times3}$, opacity $\alpha_i\in[0,1]$, and view-dependent color $\mathbf{c}_i$~\citep{3dgs}.
Following the EWA formulation~\citep{zwicker2001ewa}, a camera induces a local affine approximation of the projection at each primitive center,
\begin{equation}
\mathbf{u}(\mathbf{x}) \;\approx\; \hat{\mathbf{p}}_i + \mathbf{M}_i\,(\mathbf{x} - \boldsymbol{\mu}_i),
\qquad
\mathbf{M}_i = \mathbf{J}_i \mathbf{W} \in \mathbb{R}^{2\times 3},
\label{eq:affine}
\end{equation}
where $\mathbf{W}$ is the world-to-camera rotation, $\mathbf{J}_i$ the projection Jacobian at $\boldsymbol{\mu}_i$, and $\hat{\mathbf{p}}_i$ the projected center.
The screen-space covariance is $\mathbf{A}_i = \mathbf{M}_i \boldsymbol{\Sigma}_i \mathbf{M}_i^\top$, and the contribution of primitive $i$ to a pixel $\mathbf{p}$ is
$\alpha_i\,\mathcal{K}_G(\boldsymbol{\delta}_i;\mathbf{A}_i)$
with $\boldsymbol{\delta}_i = \hat{\mathbf{p}}_i - \mathbf{p}$ and $\mathcal{K}_G(\boldsymbol{\delta};\mathbf{A}) = \exp(-\tfrac12 \boldsymbol{\delta}^\top \mathbf{A}^{-1} \boldsymbol{\delta})$; contributions are alpha-composited front-to-back.
This per-pixel weight is, up to normalization, the integral of the 3D Gaussian density along the pixel's viewing ray under the affine approximation~\eqref{eq:affine}, an identity we exploit below.
Our operator changes this per-primitive weight but retains the base rasterizer's
depth order and front-to-back compositing.

\paragraph{Clipping plane.}
A clipping plane with unit normal $\mathbf{n}$ and offset $\tau$ keeps the half-space
$\mathcal{H}_\tau = \{\mathbf{x} \in \mathbb{R}^3 : \mathbf{n}^\top\mathbf{x} \le \tau\}$.
The reference renderer removes density outside $\mathcal{H}_\tau$ before ray
integration through its volumetric half-space clipping interface. XClipGS and
all operator ablations accept arbitrary $(\mathbf{n},\tau)$. For obliquely
acquired scans, voxel-axis planes are generally rotated and noncanonical in
world coordinates; the arbitrary-normal evaluation instead specifies planes
directly in physical-world coordinates.
Throughout, let
\begin{equation}
\sigma_n^2 = \mathbf{n}^\top \boldsymbol{\Sigma} \mathbf{n},
\quad
a = \frac{\tau - \mathbf{n}^\top\boldsymbol{\mu}}{\sigma_n},
\quad
\lambda(a) = \frac{\phi(a)}{\Phi(a)},
\label{eq:trunc_notation}
\end{equation}
denote the primitive's variance along the plane normal, its standardized signed distance to the plane, and the inverse Mills ratio, where $\phi$ and $\Phi$ are the standard normal density and CDF.
$\Phi(a)$ is precisely the fraction of the primitive's mass on the kept side.

\subsection{Half-space clipping operators}
\label{subsec:operators}

\paragraph{Clipping operators.}
A clip operator specifies how a primitive contributes when rendering under
$\mathcal{H}_\tau$.
The target primitive is the unnormalized restricted density
$r(\mathbf{x})=\mathcal{N}(\mathbf{x};\boldsymbol{\mu},\boldsymbol{\Sigma})
\mathds{1}[\mathbf{x}\in\mathcal{H}_\tau]$; our operator evaluates its ray
integral in closed form (Proposition~\ref{prop:exact}).
We contrast it with two reference points. A hard cull represents the
whole-primitive decisions used by prior work~\citep{li2025clipgs}; a
moment-matched truncation is a smooth Gaussian approximation derived from
classical truncated-normal statistics. Comparing them in one renderer isolates
the cost of binary truncation and re-Gaussianization relative to integrating the
restricted density itself. We abbreviate the hard cull, moment-matched
truncation, and our analytic operator as \opCull{}, \opMoment{}, and
\opAnalytic{}.

\paragraph{Hard cull (HC).}
This operator keeps or drops each primitive \emph{whole} using
$\mathds{1}[\mathbf{n}^{\top}\boldsymbol{\mu}\leq\tau]$.
It adds no per-pixel factor, but straddling primitives overshoot or vanish,
producing a center-quantized boundary as the plane sweeps. The raw decision is
also non-differentiable. ClipGS trains its per-primitive decision through a
straight-through estimator and deforms primitives; HC isolates the underlying
binary operator from those learned components.

\paragraph{Moment-matched truncation (MM).}
This surrogate, which we derive from classical statistics rather than take from
prior work, replaces each straddling primitive by the Gaussian whose first two
moments match the normalized truncated density $r/\Phi(a)$. Equivalently, it
minimizes $D_{\mathrm{KL}}(r/\Phi(a)\,\|\,q)$ over Gaussian $q$. The required one-sided
truncated-normal moments~\citep{tallis1961moment}, propagated through the joint
covariance, use $\mathbf{d}_n=\boldsymbol{\Sigma}\mathbf{n}/\sigma_n$ and are
\begin{equation}
\begin{aligned}
\alpha' &= \Phi(a)\,\alpha, \qquad
\boldsymbol{\mu}' = \boldsymbol{\mu}-\lambda(a)\,\mathbf{d}_n,\\
\boldsymbol{\Sigma}' &= \boldsymbol{\Sigma}+[v(a)-1]\,\mathbf{d}_n\mathbf{d}_n^\top.
\end{aligned}
\label{eq:o2}
\end{equation}
where $v(a) = 1 - a\,\lambda(a) - \lambda(a)^2$.
Opacity scales by the kept mass, the mean shifts to the truncated centroid, and the variance contracts along the clip normal by the truncated-variance factor $v(a)\in(0,1]$.
MM is smooth in $(\boldsymbol{\mu},\boldsymbol{\Sigma},\alpha,\tau)$ and can
supervise the cut face directly. Its limitation is intrinsic:
re-Gaussianization restores unbounded support, so some mass still renders beyond
the plane and the cut becomes a smooth tail of width
$\mathcal{O}(\sigma_n)$ rather than an edge.

\paragraph{Exact per-pixel integral (\opAnalytic{}).}
We render the truncated density itself rather than re-Gaussianizing it.
Under the affine approximation~\eqref{eq:affine}, screen position $\mathbf{u}$
and clip coordinate $\omega=\mathbf{n}^{\top}\mathbf{x}$ are jointly Gaussian.
Conditioning on $\mathbf{u}{=}\mathbf{p}$ factorizes the truncated ray integral
into the unclipped footprint and
$\mathrm{P}(\omega\leq\tau\mid\mathbf{u}{=}\mathbf{p})$.
With $\mathbf{b}=\mathbf{M}\boldsymbol{\Sigma}\mathbf{n}\in\mathbb{R}^{2}$,
the conditional variance is
\begin{equation}
s^2 \;=\; \sigma_n^2 - \mathbf{b}^\top \mathbf{A}^{-1} \mathbf{b},
\label{eq:cond_var}
\end{equation}
and the clipped per-pixel contribution becomes
\begin{equation}
\boxed{\;
\begin{aligned}
&\alpha\,\mathcal{K}_G(\boldsymbol{\delta};\mathbf{A})\cdot\Phi\!\big(\ell(\boldsymbol{\delta})\big),
\qquad
\ell(\boldsymbol{\delta}) = \mathbf{k}^\top\boldsymbol{\delta} + h,\\[2pt]
&\mathbf{k} = \frac{\mathbf{A}^{-1}\mathbf{b}}{s},
\qquad
h = \frac{\tau - \mathbf{n}^\top\boldsymbol{\mu}}{s}.
\end{aligned}
\;}
\label{eq:o3}
\end{equation}
Up to normalization, this Gaussian--CDF product has the classical
selection-normal form~\citep{arellanovalle2006selection}. We specialize it
to half-space EWA ray integration and derive the
rasterizer coefficients and gradients.
The clip thus reduces to three transient coefficients $(\mathbf{k}, h)$ computed
from existing geometry per primitive and view, plus one CDF evaluation per
sample with argument affine in the pixel offset $\boldsymbol{\delta}$ already
available to the rasterizer. These are render-time intermediates, not learned parameters.

\paragraph{Why the factor is per pixel.}
The marginal kept-mass fraction $\Phi(a)$ from
Equation~\ref{eq:trunc_notation} is a single scalar for the whole primitive.
Multiplying the projected footprint uniformly by this value would attenuate the
splat but could not locate the cut within it. Equation~\ref{eq:o3} instead uses
the conditional probability given the pixel. The term
$\mathbf{k}^{\top}\boldsymbol{\delta}$ moves the transition across the
footprint, while $h$ shifts it as the plane moves. Consequently, pixels whose
rays pass through the retained side approach a factor of one and those on the
removed side approach zero, with the transition width determined by the
remaining uncertainty $s$ along the ray.

\begin{proposition}[Exactness]
\label{prop:exact}
Under the local affine approximation~\eqref{eq:affine}, the
contribution~\eqref{eq:o3} equals the viewing-ray integral of the unnormalized
half-space-restricted density $r$, up to the global normalization shared with the
unclipped splat.
\end{proposition}

Conditioning the jointly Gaussian screen and clip coordinates on
$\mathbf{u}{=}\mathbf{p}$ factorizes the ray integral into the EWA footprint and
$\mathrm{P}(\omega\leq\tau\mid\mathbf{u}{=}\mathbf{p})$. Its
Schur-complement variance is Equation~\ref{eq:cond_var}, and its affine
conditional mean yields Equation~\ref{eq:o3}. Supplement~S1 gives the full
derivation.

\paragraph{Scope and boundary behavior.}
Proposition~\ref{prop:exact} is exact for one primitive's ray integral within
the affine EWA model~\eqref{eq:affine}~\citep{zwicker2001ewa} and adds no
approximation to its unclipped footprint. At image level, \xclipgs{} retains
the base rasterizer's standard center-based splat order and front-to-back alpha
compositing; the proposition isolates the added half-space integral from these
established renderer choices. A perspective-ray formulation requires
re-derived conditional coefficients~\citep{huang2026threedgeer}. Within affine
EWA, the boundary width is $s\leq\sigma_n$; as the plane becomes parallel to
the rays, $s\to0$ and
$\Phi(\ell)$ approaches a per-pixel step without MM's tail.

\paragraph{Differentiability.}
For $s>0$ our operator is smooth in primitive parameters and in $\tau$; at the
degenerate $s\to0$ limit the factor becomes a step (a plane parallel to the ray).
Since $\partial_{\ell}\Phi=\phi(\ell)$, gradients pass through
$(\mathbf{k},h)$ and the center to
$(\boldsymbol{\mu},\boldsymbol{\Sigma})$. Thus MM and Ours supervise the cut itself. HC has no gradient through its binary decision and updates only kept
primitives, the discontinuity ClipGS addresses with an STE.

\subsection{Clip-aware interior supervision}
\label{subsec:supervision}

\paragraph{Unconstrained interiors.}
A proxy trained only from external views leaves hidden density weakly
constrained: primitives behind the visible surface receive gradients only
through accumulated transmittance, and the anatomy exposed by a new cut was
never observed directly. Interior fidelity therefore depends on supervision as
well as the clip operator. The same external images can be explained by
different interior configurations that disagree once the front layers are
removed. Clipped views resolve this ambiguity by directly exposing hidden
density, not by adding
cameras around the intact outer surface alone. The analytic operator makes the
cut itself differentiable, turning interior
supervision into ordinary photometric training: gradients from clipped
reference views flow through $\Phi$ to primitives at the exposed face.

\paragraph{Training-view generation.}
For each scan, the reference volume renderer generates both intact and clipped
views across a range of camera distances. Intact views orbit and frame the whole
scan. For clipped views, we sample one of the three voxel axes and an offset
$\tau \sim \mathcal{U}[\tau_{\min}, \tau_{\max}]$ spanning the volume core, then
aim the camera obliquely at the exposed face. Each frame records its plane
$(\mathbf{n},\tau)$, and the same plane is passed to the rasterizer during
training. Varying distance supplies both full-volume context and close-up
detail; varying axis and offset exposes diverse cross-sections rather than a
single prescribed cut.

\paragraph{Training objective.}
Training minimizes the standard 3DGS photometric loss through the differentiable
clip~\citep{3dgs,wang2004ssim},
\begin{equation}
\mathcal{L} \;=\; (1-\lambda_{\mathrm{s}})\,\mathcal{L}_1\!\left(\mathcal{R}_{\mathrm{clip}}(\mathcal{G};\,\pi_v, \mathbf{n}_v, \tau_v),\, I_v\right) \;+\; \lambda_{\mathrm{s}}\,\mathcal{L}_{\mathrm{D\text{-}SSIM}},
\label{eq:loss}
\end{equation}
where $\mathcal{R}_{\mathrm{clip}}$ renders the primitive set $\mathcal{G}$ under
camera $\pi_v$ at plane $(\mathbf{n}_v,\tau_v)$ ($\tau_v{=}\infty$ for intact
views), $I_v$ is the reference render, and $\lambda_{\mathrm{s}}=0.2$ in all
experiments.
Gradients from the exposed cross-section flow through $\Phi$ into the position,
covariance, and opacity of interior primitives near the cut.
Section~\ref{sec:experiments} tests both held-out offsets along the training axes
and arbitrary plane orientations without retraining.

\subsection{Implementation}
\label{subsec:implementation}

\begin{table*}[t]
\centering
\caption{\textbf{Held-out quality and speed on eight volumes.}
FPS is measured at $1600{\times}1600$ on an A100. \textbf{Bold} marks the best.}
\label{tab:operator-grid}
\setlength{\tabcolsep}{5pt}
\resizebox{\textwidth}{!}{%
\begin{tabular}{l c@{\hskip 3pt}c@{\hskip 3pt}c@{\hskip 3pt}c @{\hskip 6pt}
                  c@{\hskip 3pt}c@{\hskip 3pt}c@{\hskip 3pt}c @{\hskip 6pt}
                  c@{\hskip 3pt}c@{\hskip 3pt}c@{\hskip 3pt}c @{\hskip 6pt}
                  c@{\hskip 3pt}c@{\hskip 3pt}c@{\hskip 3pt}c}
\toprule
& \multicolumn{4}{c}{\textbf{XClipGS (Ours)}} & \multicolumn{4}{c}{ClipGS (reimpl.)~\citeyearpar{li2025clipgs}} & \multicolumn{4}{c}{MM} & \multicolumn{4}{c}{HC} \\
\cmidrule(lr){2-5}\cmidrule(lr){6-9}\cmidrule(lr){10-13}\cmidrule(lr){14-17}
Scene & PSNR\,$\uparrow$ & SSIM\,$\uparrow$ & LPIPS\,$\downarrow$ & FPS\,$\uparrow$ & PSNR\,$\uparrow$ & SSIM\,$\uparrow$ & LPIPS\,$\downarrow$ & FPS\,$\uparrow$ & PSNR\,$\uparrow$ & SSIM\,$\uparrow$ & LPIPS\,$\downarrow$ & FPS\,$\uparrow$ & PSNR\,$\uparrow$ & SSIM\,$\uparrow$ & LPIPS\,$\downarrow$ & FPS\,$\uparrow$ \\
\midrule
\texttt{abdomen}& \textbf{33.30} & \textbf{.965} & \textbf{.083} & \textbf{682} & \ClipGSgelpsnr & \ClipGSgelssim & \ClipGSgellpips & \ClipGSgelfps & 32.97 & .964 & .084 & 615 & 33.02 & .963 & .085 & 651 \\
\texttt{intestine}& \textbf{31.77} & \textbf{.949} & \textbf{.093} & 597 & \ClipGSintestinepsnr & \ClipGSintestinessim & \ClipGSintestinelpips & \ClipGSintestinefps & 31.49 & .948 & .094 & 575 & 31.50 & .947 & .095 & \textbf{599} \\
\texttt{knee-joint}& \textbf{29.14} & \textbf{.913} & \textbf{.185} & 586 & \ClipGSkneejointpsnr & \ClipGSkneejointssim & \ClipGSkneejointlpips & \ClipGSkneejointfps & 28.85 & .912 & .186 & 580 & 28.44 & .908 & .191 & \textbf{604} \\
\texttt{lower-limb}& \textbf{34.28} & \textbf{.980} & \textbf{.026} & 670 & \ClipGSlowerpsnr & \ClipGSlowerssim & \ClipGSlowerlpips & \ClipGSlowerfps & 34.20 & .980 & .026 & 653 & 34.04 & .980 & .026 & \textbf{693} \\
\texttt{vascular}& \textbf{36.38} & \textbf{.975} & \textbf{.031} & \textbf{666} & \ClipGSvascularpsnr & \ClipGSvascularssim & \ClipGSvascularlpips & \ClipGSvascularfps & 36.18 & .975 & .031 & 621 & 36.10 & .974 & .032 & 654 \\
\texttt{heart}& \textbf{29.20} & \textbf{.940} & \textbf{.095} & \textbf{603} & \ClipGSheartpsnr & \ClipGSheartssim & \ClipGSheartlpips & \ClipGSheartfps & 28.72 & .938 & .097 & 579 & 28.65 & .936 & .101 & 594 \\
\texttt{nose}\,(MRI)& \textbf{35.92} & \textbf{.977} & \textbf{.047} & \textbf{699} & \ClipGSnosepsnr & \ClipGSnosessim & \ClipGSnoselpips & \ClipGSnosefps & 35.67 & .976 & .048 & 698 & 35.55 & .975 & .049 & 694 \\
\texttt{hand}\,(MRI)& \textbf{38.48} & \textbf{.978} & \textbf{.036} & 740 & \ClipGShandpsnr & \ClipGShandssim & \ClipGShandlpips & \ClipGShandfps & 38.22 & .977 & .037 & 724 & 37.85 & .975 & .040 & \textbf{772} \\
\midrule
\texttt{avg}& \textbf{33.56} & \textbf{.960} & \textbf{.075} & 655 & \ClipGSpsnr & \ClipGSssim & \ClipGSlpips & \ClipGSfps & 33.29 & .959 & .075 & 631 & 33.14 & .957 & .077 & \textbf{658} \\
\bottomrule
\end{tabular}}
\vspace{-1em}
\end{table*}

\paragraph{Rasterization.}
Our CUDA implementation combines Speedy-Splat
culling~\citep{hanson2024speedy}, TC-GS~\citep{tcgs2025}, and Mip-Splatting's
3D filter~\citep{yu2023mipsplattingaliasfree3dgaussian}. Preprocessing computes
the shared truncation statistics. MM then rewrites each straddling primitive
once, whereas Ours retains its ordinary EWA footprint and evaluates
$\Phi(\ell)$ only for samples from straddling primitives. Both paths remain
differentiable end to end and accept arbitrary plane orientations.
Supplements~S2--S3 give numerical safeguards, gradient validation, and
experimental configurations.

\section{Experiments}
\label{sec:experiments}

We evaluate held-out plane offsets and orientations, cut-face errors, and
render-time operator swaps. Ours, MM, and HC share the \dgs{} backbone,
initialization, views, and planes; ClipGS uses the same initialization, views,
planes, and iteration budget with its 3DGS representation.

\subsection{Experimental protocol}
\label{subsec:protocol}

\paragraph{Datasets and splits.}
The benchmark comprises eight clinical volumes: six CT (\texttt{abdomen},
\texttt{intestine}, \texttt{knee-joint}, \texttt{lower-limb},
\texttt{heart}, \texttt{vascular}) and two MRI (\texttt{nose}, \texttt{hand}),
spanning smooth and high-frequency anatomy. A physically based Monte Carlo
volume path tracer produces a \textbf{general} set of $900$ views
($810$ train/$90$ test). Camera distance varies; half of the views are intact
and half contain a plane sampled across the volume core. Test-plane offsets are
numerically disjoint from training and drawn from the same interval, evaluating
interpolation to held-out cuts throughout the sampled volume core. A disjoint
voxel-axis \textbf{cut-eval} set uses three
fixed planes per volume and $30$ cameras, each rendered both with and without
clipping, yielding the paired references used by the localized metrics. A
second set evaluates arbitrary-normal center planes without retraining.
Supplement~S3
specifies volume preparation, training-view sampling, and rendering;
Supplement~S5 specifies both cut-eval plane sets. The general set
mixes fit-to-frame views with close-ups: intact views supply the outer anatomy
and whole-volume context, while clipped cameras target the exposed face from
oblique angles. The train/test split is stratified by view mode and plane axis,
so each regime is represented without reusing an exact offset.

\paragraph{Comparison methods.}
\textbf{Ours} is the analytic half-space clip of
Section~\ref{subsec:operators}. \textbf{MM} (forward-KL-optimal Gaussian
surrogate) and \textbf{HC} (bare per-primitive hard cull) share Ours'
\mbox{N-DGS}
backbone, initialization, planes, and supervision, isolating the operator
within one representation. \mbox{HC corresponds} to the binary operator class without
ClipGS's STE training or deformation network.
\textbf{ClipGS}~\citep{li2025clipgs} is our best-effort system-level
reimplementation from its published description, using its vanilla-3DGS
backbone, trainable visibility, and plane-conditioned deformation MLP; no public
code was available.
\textbf{RaRa}~\citep{li2025rara} enters only as a render-time operator on fixed
interiors (Section~\ref{subsec:opswap}). Supplement~S4 details both baselines.
Ours, MM, HC, and ClipGS are trained for $30{,}000$ iterations, and every
result uses the fixed final checkpoint.

\paragraph{Evaluation metrics.}
On general views we report PSNR, SSIM~\citep{wang2004ssim},
LPIPS~\citep{zhang2018lpips}, and FPS. Cut-eval adds four
cut-face metrics. \bandS{} is band SSIM over the exposed cut face. The
\emph{difference-referenced cut error} \cdem{} measures \emph{where} the cut
acts: for the method and reference, we subtract the clipped render from its
unclipped companion, threshold the resulting removal map, and normalize the
symmetric-difference area by the reference region. This construction cancels
reconstruction error common to both renders and penalizes both missed cuts and
over-removal. We report it over the exposed face (\cdemP) and grazing edge
(\cdemG). \leakm{} is the fraction of near-plane foreground energy on the culled
side, where the reference is empty. As an image-space reference metric, it
measures agreement with the independent path-traced reference and also captures
proxy and renderer mismatch.
Supplement~S5 gives the equations, per-volume results, parameter sensitivity,
and the camera-independent wrong-side-mass diagnostic
CErr\textsubscript{3D}, which measures operator geometry and is zero for Ours
by analytic definition. Band PSNR differs by only hundredths and is omitted.

\subsection{Held-out quality and speed}
\label{subsec:global}

\begin{figure*}[t]
\centering
\includegraphics[width=\textwidth]{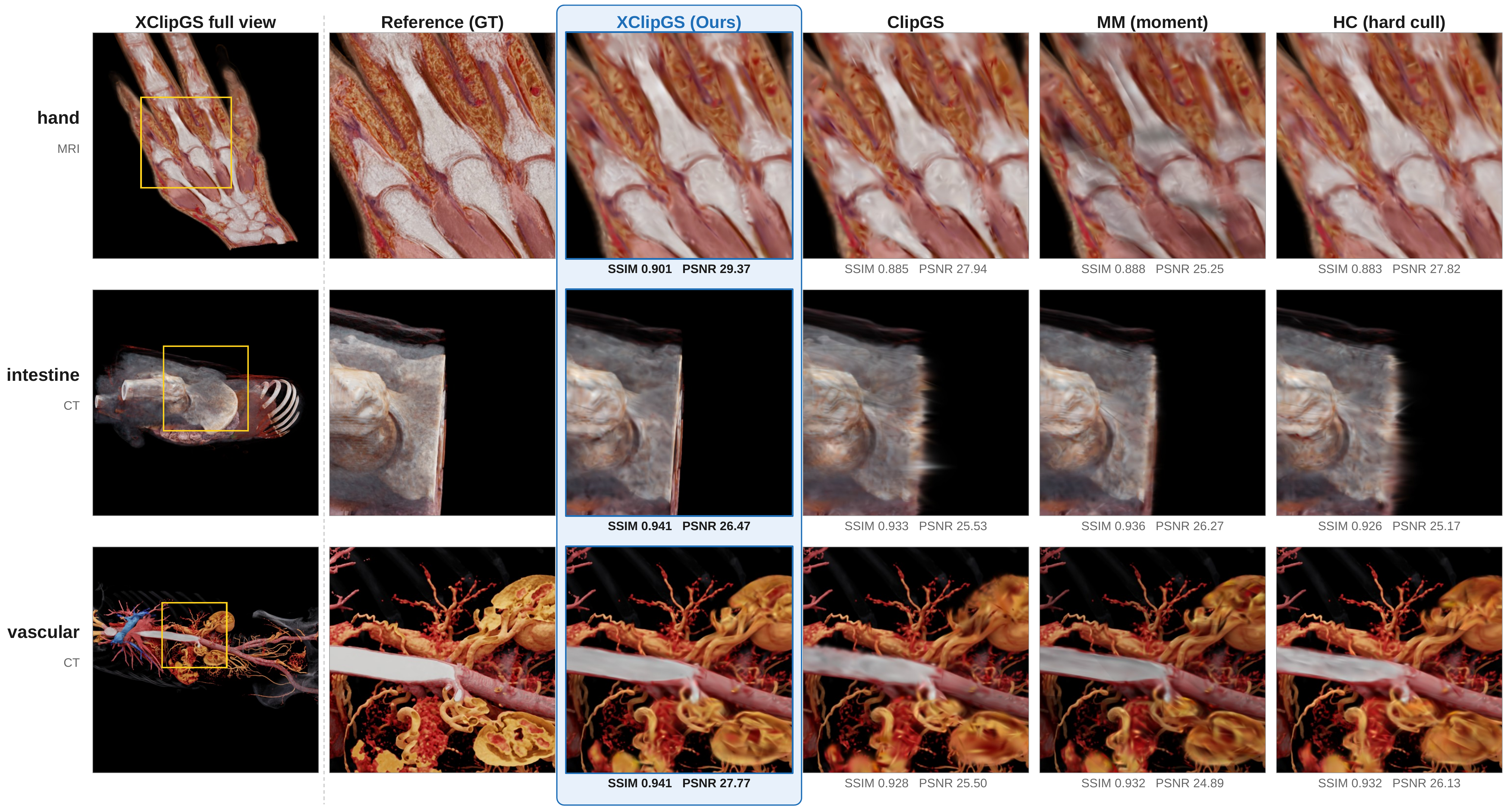}
\caption{\textbf{Held-out qualitative comparison.}
Full views locate each crop; the middle row is an unseen arbitrary-normal
grazing view. The remaining columns compare matched reference and method crops,
with SSIM and PSNR shown below.}
\label{fig:qualitative}
\vspace{-1.3em}
\end{figure*}

\paragraph{Image quality.}
Table~\ref{tab:operator-grid} reports held-out quality and speed. Our method has
the highest PSNR on every volume and improves over the ClipGS reimplementation
by \ClipGSgap{} on average PSNR. This is a system-level comparison because
ClipGS uses the vanilla 3DGS representation specified in its paper. The controlled \dgs{}
ablations trace the operator effect: the
smooth surrogate (MM, $33.29$~dB) and the bare hard cull (HC, $33.14$) both trail
Ours ($33.56$), with the same overall trend in SSIM and LPIPS. Because every
evaluation offset is disjoint from training, these scores show that the trained
system maintains fidelity when interpolating to numerically held-out
cross-sections along the sampled axes. The advantage also holds across all
eight volumes rather than being driven by one scan: Ours leads PSNR in every
row, from smooth soft tissue to thin vessels and fine joint structure. Global
metrics nevertheless
dilute the operator effect, because half of the held-out views contain no cut and,
in the remaining views, the boundary occupies only a small image region. The largest
Ours--HC gap occurs on \texttt{knee-joint} ($+0.70$~dB), which contains fine
cut-face structure. This dilution motivates the localized evaluation in
Section~\ref{subsec:cutface}.

\paragraph{Rendering speed.}
The analytic cut adds little measured overhead: Ours renders at $655$~FPS,
matching the hard
cull (HC, $658$) that does no per-sample work and the moment-matched surrogate
(MM, $631$), with all three within a few percent. Thus, the closed-form
half-space factor adds negligible cost at this resolution on the measured A100.
The ClipGS
reimplementation runs at \ClipGSfps~FPS ($\sim\!2.4\times$ slower), reflecting its
extra plane-conditioned network per frame. The controlled Ours/MM/HC comparison
shows the analytic operator itself runs at approximately hard-cull speed rather
than trading fidelity for interactivity.

\subsection{Cut-face fidelity and localization}
\label{subsec:cutface}
\paragraph{Cut-eval protocol.}
To measure the clip where the global metric dilutes it, the cut-face metrics act on
the held-out cut-eval set: three fixed planes per volume (one per voxel axis) at
the volume core, each viewed by five \emph{perpendicular} cameras (looking down the
normal, so the cut face fills the frame) and five \emph{grazing} cameras (tilted
$1.5^\circ$ off edge-on, exposing the cut edge). The clipped ground truth is
rendered by the reference volume renderer with the plane applied by the renderer
itself, and each camera also renders an \emph{unclipped} companion for the
difference-referenced error, supplying a matched volumetric reference for the
partial-clipping behavior prior work calls ``hard to
define''~\citep{li2025rara} (details in Supplement~S5).
To test orientation generalization, a second set samples five arbitrary physical-
world normals per volume, each at least $20^\circ$ from every voxel axis and
from the other sampled normals. Each
plane passes through the volume center and is viewed once perpendicularly and
twice at grazing incidence, giving $15$ views per volume. We evaluate the same
checkpoints without retraining.

\begin{table*}[t]
\centering
\captionsetup{skip=2pt}
\caption{\textbf{Average cut-face results over eight volumes.}
\bandS: band SSIM; \cdemG/\cdemP: edge/face error; \leakm: leakage. Errors are
$\times10^{-2}$; \textbf{bold}: best per row.}
\label{tab:cutface}
\setlength{\tabcolsep}{3pt}
\resizebox{\textwidth}{!}{%
\begin{tabular}{l rccc @{\hskip 6pt} rccc @{\hskip 6pt} rccc @{\hskip 6pt} rccc}
\toprule
& \multicolumn{4}{c}{\textbf{XClipGS (Ours)}} & \multicolumn{4}{c}{ClipGS (reimpl.)~\citeyearpar{li2025clipgs}} & \multicolumn{4}{c}{MM} & \multicolumn{4}{c}{HC} \\
\cmidrule(lr){2-5}\cmidrule(lr){6-9}\cmidrule(lr){10-13}\cmidrule(lr){14-17}
Plane set & \bandS\,$\uparrow$ & \cdemG\,$\downarrow$ & \cdemP\,$\downarrow$ & \leakm\,$\downarrow$ & \bandS\,$\uparrow$ & \cdemG\,$\downarrow$ & \cdemP\,$\downarrow$ & \leakm\,$\downarrow$ & \bandS\,$\uparrow$ & \cdemG\,$\downarrow$ & \cdemP\,$\downarrow$ & \leakm\,$\downarrow$ & \bandS\,$\uparrow$ & \cdemG\,$\downarrow$ & \cdemP\,$\downarrow$ & \leakm\,$\downarrow$ \\
\midrule
Voxel-axis & \textbf{.860} & \textbf{22.7} & \textbf{17.9} & \textbf{0.40} & .809 & 33.4 & 19.5 & 16.42 & .853 & 23.6 & 18.4 & 6.18 & .819 & 29.4 & 18.5 & 12.54 \\
Arbitrary-normal & \textbf{.874} & \textbf{23.5} & \textbf{14.3} & \textbf{0.13} & .844 & 35.9 & 15.7 & 17.24 & .870 & 25.5 & 15.1 & 5.67 & .842 & 35.9 & 14.9 & 18.30 \\
\bottomrule
\end{tabular}}
\vspace{-0.5em}
\end{table*}

\begin{table*}[t]
\centering
\captionsetup{skip=2pt}
\caption{\textbf{Fixed-interior operator swap.}
Each row changes only the clip rule on the voxel-axis set; metrics follow
Table~\ref{tab:cutface}. \textbf{Bold}: full-precision row best; per-volume:
Supplement~S7.}
\label{tab:opswap}
\setlength{\tabcolsep}{3pt}
\resizebox{\textwidth}{!}{%
\begin{tabular}{l rccc @{\hskip 6pt} rccc @{\hskip 6pt} rccc @{\hskip 6pt} rccc}
\toprule
& \multicolumn{4}{c}{\textbf{Ours (analytic)}} & \multicolumn{4}{c}{RaRa~\citeyearpar{li2025rara}} & \multicolumn{4}{c}{MM} & \multicolumn{4}{c}{HC} \\
\cmidrule(lr){2-5}\cmidrule(lr){6-9}\cmidrule(lr){10-13}\cmidrule(lr){14-17}
Fixed interior & \bandS\,$\uparrow$ & \cdemG\,$\downarrow$ & \cdemP\,$\downarrow$ & \leakm\,$\downarrow$ & \bandS\,$\uparrow$ & \cdemG\,$\downarrow$ & \cdemP\,$\downarrow$ & \leakm\,$\downarrow$ & \bandS\,$\uparrow$ & \cdemG\,$\downarrow$ & \cdemP\,$\downarrow$ & \leakm\,$\downarrow$ & \bandS\,$\uparrow$ & \cdemG\,$\downarrow$ & \cdemP\,$\downarrow$ & \leakm\,$\downarrow$ \\
\midrule
Ours-trained & \textbf{.860} & \textbf{22.7} & \textbf{17.9} & \textbf{0.40} & .856 & 23.4 & 18.3 & 6.24 & .848 & 23.7 & 18.0 & 5.85 & .838 & 30.1 & 18.2 & 14.89 \\
HC-trained & \textbf{.828} & \textbf{23.1} & 18.5 & \textbf{0.37} & .826 & 23.6 & 18.8 & 4.89 & \textbf{.828} & 23.5 & \textbf{18.4} & 5.12 & .819 & 29.4 & 18.5 & 12.54 \\
\bottomrule
\end{tabular}}
\vspace{-1.2em}
\end{table*}

\paragraph{Image-space results.}
Table~\ref{tab:cutface} and Figure~\ref{fig:qualitative} expose differences
diluted by global metrics. XClipGS has the best average on all four metrics for
both plane sets and the highest voxel-axis band SSIM on seven volumes. Without
retraining, its arbitrary-normal averages are $.874$ \bandS{}, $23.5$ \cdemG{},
$14.3$ \cdemP{}, and $0.13\times10^{-2}$ Leak. It also has the highest
displayed SSIM and PSNR across the examples, which span fine MRI structure in
\texttt{hand}, layered soft tissue in \texttt{intestine}, and thin vessels in
\texttt{vascular}. XClipGS preserves their structure and contrast while
avoiding the diffuse or jagged boundaries of the approximations.
Supplement~S5 gives the per-volume results and arbitrary-normal
CErr\textsubscript{3D}.

Because these orientations and camera-plane configurations never appear in
training, this set tests whether axis-supervised interiors support new
center-plane orientations. The preserved ordering also shows that the analytic
operator is not tied to voxel axes.

On voxel-axis cuts, Ours reduces \cdemG{} to $22.7$, versus $23.6$ for MM,
$29.4$ for HC, and $33.4$ for ClipGS; Leak falls to $0.40\times10^{-2}$, versus
$6.18$, $12.54$, and $16.42$. MM retains an unbounded tail, while HC keeps or
removes whole straddling splats. Supplement~S5 confirms the lowest face CDE at
all four thresholds and Leak in all $27$ settings; Ours wins edge CDE in $11$
of $16$ settings and MM in five.

\paragraph{Global versus geometric fidelity.}
MM nearly matches Ours in global PSNR and SSIM, yet its unbounded tail produces
about $15.5\times$ more leakage; HC is inexpensive but raises edge error. Our
factor preserves the ordinary footprint and changes only the half-space mass,
explaining its fidelity, speed, and cleaner boundary. Exactness is
per-splat and operator-level: Equation~\ref{eq:o3} integrates each represented
Gaussian inside the half-space, while projection, ordering, and compositing
remain inherited from the base renderer. CErr\textsubscript{3D} is zero by
construction even for arbitrary normals.
Image-space Leak instead compares the learned proxy with an independent
path-traced reference, so its small residual reflects density, color, and
compositing mismatch rather than outside-plane mass from our operator.

\subsection{Fixed-interior operator comparison}
\label{subsec:opswap}
\paragraph{Operator-swap protocol.}
We render fixed checkpoints through each rule on the voxel-axis set, decoupling
operator and interior. This admits RaRa~\citep{li2025rara}, whose released rule
lacks a backward path; its renderer agrees with ours to $42.8$~dB on unclipped
views, indicating small off-plane differences. The jointly trained ClipGS cull
and deformation network remain in the system-level comparison above, where the
complete learned system is evaluated.

\paragraph{Operator-swap results.}
On the Ours-trained interior, our operator has the best average
\cdemG{}/\cdemP{} ($22.7/17.9$) and lowest Leak ($0.40$, versus $6.24$ for
RaRa, $5.85$ for MM, and $14.89\times10^{-2}$ for HC;
Table~\ref{tab:opswap}). RaRa nearly matches band quality but leaks
about $16\times$ more. This agrees with the analytic distinction: a chord fraction
is not the Gaussian mass retained along the ray, whereas
Equation~\ref{eq:o3} evaluates that mass in closed form at every covered pixel.

Table~\ref{tab:cutface} additionally evaluates MM and HC on their co-adapted
interiors, while the fixed-interior test supplies RaRa with the same clip-aware
interior as the other rules. Repeating the swap on the HC-trained checkpoint
preserves Ours' average \cdemG{} and Leak lead across both interiors.
Supplement~S7 gives per-volume results;
Supplement~S4 details renderer agreement.

\subsection{Clip-aware supervision ablation}
\label{subsec:extonly}
We fix the clip rule and vary only supervision. The \emph{external-view-only}
control shares the backbone, initialization, test split, and $30{,}000$-iteration
budget but trains only on intact views. It gains $0.7$~dB on intact views yet
loses $8.3$~dB on every volume's clipped views ($32.0\!\to\!23.8$).
Voxel-axis band SSIM falls from $.860$ to $.346$ and \cdemP{} rises from $17.9$
to $25.6\times10^{-2}$. The $8.3$~dB gain from clip-aware supervision therefore
quantifies its contribution to interior fidelity.

Together, the two controls isolate complementary effects: the operator swap
attributes lower leakage to analytic clipping, while the supervision ablation
holds that operator fixed and attributes interior fidelity to clipped views.
Faithful clipping needs both (Supplement~S6, Table~S5).

\subsection{Discussion}
\xclipgs{} separates responsibilities: clip-aware supervision learns the
exposed anatomy, while the analytic renderer enforces each requested half-space.
At deployment, a plane changes only transient per-primitive coefficients and
one conditional-CDF factor per covered sample; one checkpoint therefore
supports moving voxel-axis or oblique cuts without a plane-conditioned network
or retraining. This separation also keeps the viewer interface simple: intact
and clipped rendering share one learned volume and appearance model while the
user updates only the requested plane.

\section{Conclusion}
\label{sec:conclusion}
\xclipgs{} couples an exact closed-form per-splat affine-EWA integral with
photometric interior supervision through the same differentiable operator,
without learned clipping parameters or an auxiliary network. Across eight CT
and MRI volumes, it leads all four cut-face metrics on voxel-axis and
arbitrary-normal planes at hard-cull speed. The $8.3$~dB advantage from clipped
training views confirms that faithful clipping benefits from both analytic
truncation and a supervised interior. Paired clipped/unclipped evaluation
localizes boundary error, while arbitrary-normal tests demonstrate transfer
beyond the supervised axes. Because new half-spaces require no retraining,
\xclipgs{} provides a reusable primitive for interactive Gaussian volume
viewers.

\clearpage
\bibliography{references}

\clearpage
\beginsupplement
\setcounter{section}{0}
\renewcommand{\thesection}{S\arabic{section}}
\renewcommand{\thesubsection}{\thesection.\arabic{subsection}}
\setcounter{equation}{0}
\renewcommand{\theequation}{S\arabic{equation}}
\twocolumn[{
  \centering
  {\Large\bfseries Supplementary Material}\\[0.5em]
  {\large \xclipgs{}: Exact Half-Space Clipping for Medical Volume Gaussian Splatting}
  \par\vspace{1.4em}
}]

\noindent
This supplement collects the derivations, protocol, and per-scan results that the
main paper refers to by section number. Section~\ref{supp:proof} gives the full
proof of the exactness proposition; Section~\ref{supp:impl} details the CUDA
rasterizer forward/backward and its numerical validation; Section~\ref{supp:repro} specifies
the benchmark volumes, view sampling, and training and timing setup;
Section~\ref{supp:baselines} describes the ClipGS and RaRa baseline
implementations; Section~\ref{supp:metrics} defines the cut-face metrics
(\bandS, \cdem, \leakm, \cerrD) and tests their parameter sensitivity;
Section~\ref{supp:extonly} reports
the external-view-only ablation per scan; Section~\ref{supp:fairness} gives
the operator-swap controls and per-volume tables; and
Section~\ref{supp:demos} describes the video and interactive
demonstrations available on the project page. All numbers are produced
by the same pipeline as the main paper and reuse its notation.

\section{Proof of Analytic Exactness}
\label{supp:proof}
We give the full proof of Proposition 1 in the main paper. Fix a primitive
$(\boldsymbol{\mu},\boldsymbol{\Sigma},\alpha)$ with
$\boldsymbol{\Sigma}\succ0$ and a camera with the local affine model
$\mathbf{u}(\mathbf{x})=\hat{\mathbf{p}}+\mathbf{M}(\mathbf{x}-\boldsymbol{\mu})$,
$\mathbf{M}\in\mathbb{R}^{2\times3}$ of rank $2$, and let
$\omega(\mathbf{x})=\mathbf{n}^{\top}\mathbf{x}$ be the clip coordinate.

\paragraph{An affine chart adapted to the rays.}
Under the affine model, the viewing ray of pixel $\mathbf{p}$ is the line
$\{\mathbf{x}:\mathbf{u}(\mathbf{x})=\mathbf{p}\}$, whose direction spans the
one-dimensional kernel of $\mathbf{M}$. Choose a unit vector $\mathbf{d}$ with
$\mathbf{M}\mathbf{d}=\mathbf{0}$ and set
$t(\mathbf{x})=\mathbf{d}^{\top}(\mathbf{x}-\boldsymbol{\mu})$. The map
$T:\mathbf{x}\mapsto(\mathbf{u}(\mathbf{x}),t(\mathbf{x}))$ is affine, and it is
invertible: if $\mathbf{M}\mathbf{v}=\mathbf{0}$ then
$\mathbf{v}=c\,\mathbf{d}$, and $\mathbf{d}^{\top}\mathbf{v}=c=0$. Its Jacobian
$J=\lvert\det[\mathbf{M};\mathbf{d}^{\top}]\rvert$ is a nonzero constant. Hence
if $\mathbf{x}\sim\mathcal{N}(\boldsymbol{\mu},\boldsymbol{\Sigma})$ with density
$\rho$, the pair $(\mathbf{u},t)=T(\mathbf{x})$ is jointly Gaussian with density
$f(\mathbf{u},t)=J^{-1}\rho\!\left(T^{-1}(\mathbf{u},t)\right)$, and
$\omega$, being another affine image of $\mathbf{x}$, is jointly Gaussian with
$(\mathbf{u},t)$.

\paragraph{Factorizing the clipped ray integral.}
The reference contribution of the primitive to pixel $\mathbf{p}$ under the
half-space restriction is the ray integral of the restricted density,
\begin{equation}
I_{\mathrm{clip}}(\mathbf{p})
=\alpha\!\int_{\mathbb{R}}\rho\!\left(\mathbf{x}(\mathbf{p},t)\right)
\mathds{1}\!\left[\omega(\mathbf{x}(\mathbf{p},t))\leq\tau\right]dt ,
\end{equation}
where $\mathbf{x}(\mathbf{p},t)=T^{-1}(\mathbf{p},t)$ traverses the ray. Writing
the integrand through $f$ and factorizing the joint density into marginal and
conditional, $f(\mathbf{p},t)=f_{U}(\mathbf{p})\,f_{T\mid U}(t\mid\mathbf{p})$,
\begin{align}
I_{\mathrm{clip}}(\mathbf{p})
&=\alpha J\, f_{U}(\mathbf{p})\!\int_{\mathbb{R}}
f_{T\mid U}(t\mid\mathbf{p})\,
\mathds{1}\!\left[\omega\leq\tau\right]dt\\
&=\alpha J\, f_{U}(\mathbf{p})\,
\mathrm{P}\!\left(\omega\leq\tau\mid\mathbf{u}=\mathbf{p}\right),
\end{align}
because, given $\mathbf{u}=\mathbf{p}$, the clip coordinate $\omega$ is an
affine function of $t$ alone, so integrating the conditional density of $t$
over the event $\{\omega\leq\tau\}$ is exactly its conditional probability.
Setting $\tau=\infty$ recovers the unclipped integral
$I_{\mathrm{full}}(\mathbf{p})=\alpha J f_{U}(\mathbf{p})$, so
\begin{equation}
I_{\mathrm{clip}}(\mathbf{p})
=I_{\mathrm{full}}(\mathbf{p})\cdot
\mathrm{P}\!\left(\omega\leq\tau\mid\mathbf{u}=\mathbf{p}\right).
\label{eq:supp-ratio}
\end{equation}
The marginal $f_{U}$ is the Gaussian
$\mathcal{N}(\mathbf{p};\hat{\mathbf{p}},\mathbf{A})$ with
$\mathbf{A}=\mathbf{M}\boldsymbol{\Sigma}\mathbf{M}^{\top}$, which is positive
definite because $\boldsymbol{\Sigma}\succ0$ and $\mathbf{M}$ has full row rank.
Up to the constant $2\pi\lvert\mathbf{A}\rvert^{1/2}$ this is the unnormalized
EWA footprint $\mathcal{K}_G(\boldsymbol{\delta};\mathbf{A})$ with
$\boldsymbol{\delta}=\hat{\mathbf{p}}-\mathbf{p}$; the splatting kernel absorbs
this constant and $J$ identically in the clipped and unclipped cases, which is
the ``global normalization shared with the unclipped splat'' in the proposition.
Note also that the result is independent of the choice of $\mathbf{d}$: any
other affine ray coordinate rescales $J$ and $f_{T\mid U}$ jointly and leaves
Equation~\eqref{eq:supp-ratio} unchanged.

\paragraph{The conditional law of the clip coordinate.}
The pair $(\omega,\mathbf{u})$ is jointly Gaussian with
\begin{align}
\mathrm{E}[\omega]&=\mathbf{n}^{\top}\boldsymbol{\mu},\qquad
\mathrm{Var}(\omega)=\mathbf{n}^{\top}\boldsymbol{\Sigma}\mathbf{n}=\sigma_n^2,\\
\mathrm{Cov}(\omega,\mathbf{u})&=\mathbf{n}^{\top}\boldsymbol{\Sigma}\mathbf{M}^{\top}=\mathbf{b}^{\top},
\end{align}
where $\mathbf{b}=\mathbf{M}\boldsymbol{\Sigma}\mathbf{n}$. The standard
Gaussian conditioning formulas give
\begin{align}
\mathrm{E}[\omega\mid\mathbf{u}=\mathbf{p}]
&=\mathbf{n}^{\top}\boldsymbol{\mu}
+\mathbf{b}^{\top}\mathbf{A}^{-1}(\mathbf{p}-\hat{\mathbf{p}})\nonumber\\
&=\mathbf{n}^{\top}\boldsymbol{\mu}
-\mathbf{b}^{\top}\mathbf{A}^{-1}\boldsymbol{\delta},\\
\mathrm{Var}(\omega\mid\mathbf{u}=\mathbf{p})
&=\sigma_n^{2}-\mathbf{b}^{\top}\mathbf{A}^{-1}\mathbf{b}\;=\;s^{2},
\end{align}
the Schur complement of $\mathbf{A}$ in the joint covariance of
$(\omega,\mathbf{u})$. For $s>0$, standardizing yields
\begin{align}
\mathrm{P}\!\left(\omega\leq\tau\mid\mathbf{u}=\mathbf{p}\right)
&=\Phi\!\left(\frac{\tau-\mathbf{n}^{\top}\boldsymbol{\mu}
+\mathbf{b}^{\top}\mathbf{A}^{-1}\boldsymbol{\delta}}{s}\right)\nonumber\\
&=\Phi\!\left(\mathbf{k}^{\top}\boldsymbol{\delta}+h\right),
\end{align}
with $\mathbf{k}=\mathbf{A}^{-1}\mathbf{b}/s$ and
$h=(\tau-\mathbf{n}^{\top}\boldsymbol{\mu})/s$, using the symmetry of
$\mathbf{A}^{-1}$. Substituting into Equation~\eqref{eq:supp-ratio} gives the
clipped per-pixel contribution
$\alpha\,\mathcal{K}_G(\boldsymbol{\delta};\mathbf{A})\,
\Phi(\mathbf{k}^{\top}\boldsymbol{\delta}+h)$, which is Equation~\ref{eq:o3} of the main
paper. \hfill$\blacksquare$

\paragraph{Degenerate cases.}
$s^{2}\geq0$ always, with $s^{2}=0$ exactly when $\omega$ is almost surely an
affine function of $\mathbf{u}$, i.e., when the clip coordinate is determined by
the pixel; geometrically the plane then contains the viewing-ray direction. In
that limit the conditional law collapses to a point mass and
$\Phi(\ell)\to\mathds{1}[\,\mathrm{E}[\omega\mid\mathbf{u}]\leq\tau\,]$, the
per-pixel step described in the main paper. The finite-precision kernel handles
this regime with the conditional-variance floor documented in
Section~\ref{supp:impl}.

\paragraph{The operator ladder.}
In the same sense, the three operators of the main paper render ray integrals of
the same restricted density $r$ under successively weaker projections of it: MM
renders the ray integral of the Gaussian moment projection of $r$, and HC the
ray integral of $r$'s binary keep/drop projection. Comparing them therefore
isolates operator fidelity on a single ladder.

\paragraph{Scope of exactness.}
The proof is per primitive: it establishes the exact retained ray integral
relative to that primitive's affine-EWA footprint. The complete rasterizer
continues to use the base renderer's center-based splat order and front-to-back
alpha compositing. The proposition thus isolates the added half-space integral,
while the complete image retains the established ordering and compositing
approximations of the base EWA rasterizer.

\section{Rasterizer Implementation and Validation}
\label{supp:impl}
\paragraph{Forward pass.}
HC, MM, and Ours share a tile-based CUDA rasterizer that combines Speedy-Splat's
tight tile culling~\citep{hanson2024speedy}, the packed half-precision
tensor-core forward of TC-GS~\citep{tcgs2025}, and the Mip-Splatting 3D low-pass
filter~\citep{yu2023mipsplattingaliasfree3dgaussian}. Before rasterization, MM
and Ours discard primitives with negligible kept mass $\Phi(a)\leq10^{-6}$. MM
also applies the $1/255$ compositing threshold to its mass-scaled opacity
$\alpha\Phi(a)$. MM rewrites
$(\boldsymbol{\mu},\boldsymbol{\Sigma},\alpha)$ during preprocess, whereas Ours
computes the three clip coefficients and multiplies sample opacity by
$\Phi(\ell)$ only for plane-straddling primitives. To avoid division by zero
near the degenerate limit, the implementation uses
$s_{\mathrm{impl}}^2=\max(s^2,10^{-8})$ and evaluates the CDF in fp32. The
backward pass sets the derivative through $s^2$ to zero while this floor is
active; this branch is evaluated only for clipped scenes. Because $\ell$ is affine in
the pixel coordinate, the factor integrates into the packed forward with its
coefficients staged per tile. The existing forward path is retained, with one
scalar fp32 CDF evaluation added for each straddling sample.
Here ``straddling'' names the finite-precision rasterizer's active clipping
branch. The $10^{-6}$ retained-mass cutoff and $10^{-8}$ variance floor provide
its numerical safeguards, while Proposition 1 describes the corresponding
ideal-Gaussian operator.

\paragraph{Backward pass.}
The backward pass emits gradients for the clip coefficients and projected center
and chains them through the conic and projection Jacobian to
$(\boldsymbol{\mu},\boldsymbol{\Sigma})$. When the variance floor is active, its
derivative with respect to the unclamped $s^2$ is set to zero; all other paths
retain their finite analytic gradients. Training uses a warp-per-splat render
backward in the style of Faster-GS~\citep{hahlbohm2026fastergs}, in which each
lane accumulates one primitive's full gradient, including its three clip
coefficients, in registers and issues a single atomic per primitive per tile;
on an A100 at $800{\times}800$ with $300$k Gaussians and an active oblique clip,
this reduces a measured forward--backward step from $5.47$ to $3.50$~ms
($1.6\times$) relative to the legacy per-pixel backward.

\paragraph{Numerical validation and overhead.}
We validate forward and backward passes against a float64 autograd
implementation of the full pipeline. All parameter gradients match to
approximately $10^{-6}$ relative error for axis-aligned and oblique planes, with
clipping active and inactive. A separate robustness suite covers rotated camera
poses, partial tiles at image borders, alpha saturation, early ray termination,
and rotation equivariance of the clip (rotating the scene and the plane together
leaves the render unchanged). Relative to MM, Ours requires one transient
$16$-byte per-view buffer per primitive and one CDF evaluation per straddling
sample; it introduces no learned primitive parameters.
These float64 tests provide independent numerical validation of the implemented
formulas and safeguards; Section~\ref{supp:metrics}'s zero-by-construction
diagnostic instead measures operator geometry.

\section{Benchmark and Training Setup}
\label{supp:repro}
\paragraph{Volumes and reference renderer.}
The eight volumes are de-identified clinical CT and MRI scans; \texttt{nose} and
\texttt{hand} are MRI, the other six are CT. Each is read from its DICOM series
and rendered on its own voxel grid (for example, \texttt{heart} is
$512\times512\times317$), with volumes kept on their native grids rather than
resampled to a common resolution. For each scan, a fixed render preset specifies
the intensity-to-RGBA transfer function, material parameters, environment and
area lights, and Monte Carlo quality settings for every view. All presets use
the \texttt{uffizi-large.hdr} environment map; their scan-specific progressive
rendering limits range from $1{,}000$ to $7{,}135$ iterations. The cinematic,
physically based path tracer is the reference throughout: all
``ground truth'' and clipped reference images are its output. Clipping uses its
native volumetric half-space interface rather than a splatting operator. Thus
clipped and unclipped renders from one camera share the volume, transfer
function, lighting, materials, and camera and differ only by the half-space
clip. The exact
per-volume metadata, render preset, and renderer configuration will be released with
the code. The benchmark measures rendering fidelity on these eight scans, each
treated as one scene.

\paragraph{Views and sampling.}
Every image is rendered at $1600\times1600$ with a $60^{\circ}$ vertical field
of view. We generate each $900$-view set once with random seed $7$: $300$ intact
fit-to-frame views at distance factor $1.0$, $300$ clipped near-fit views at
$[0.85,1.0]$, $150$ clipped close-ups at $[0.4,0.6]$, and $150$ intact
close-ups at $[0.4,0.6]$. For a clipped view, we sample one of the three voxel
axes uniformly and an offset $\tau$ uniformly over the central $30\%$--$70\%$
of its spatial extent. The camera targets a point on the exposed face and is
sampled $25^{\circ}$--$70^{\circ}$ from the plane normal. The $810/90$
train/test split is stratified within every view-mode and cut-axis combination;
conversion preserves this split rather than sampling it again. Test offsets are
therefore numerically disjoint from training offsets but come from the same
offset interval, measuring interpolation to held-out cuts throughout the
sampled volume core. The voxel-axis cut-evaluation set
contains $30$ cameras per volume, each producing clipped and unclipped renders
on three fixed planes (one per voxel axis) at the volume-core midpoint. A
separate orientation set uses five deterministic physical-world center planes
per volume, with one perpendicular and two grazing views per plane. Both
evaluation sets are held out from optimization, and the arbitrary-normal set
uses the original checkpoints without retraining.

\paragraph{Coordinate alignment.}
After reference rendering, camera poses, clipping planes, and initialization
points are transformed into one volume-centered physical coordinate frame. Each
cut is stored as the retained half-space
$\mathbf{n}^{\top}\mathbf{x}\leq\tau$. For an obliquely oriented DICOM grid,
the selected voxel-axis normal and its offset are transformed into patient/world
coordinates; the volume itself remains on its native grid. For the
arbitrary-normal set, we instead sample a physical-world normal and center
plane and convert it to the renderer's index-coordinate plane equation. The
initialization point cloud is centered once by the same volume-center
translation, ensuring that the images, cameras, planes, and primitives share
one frame. In both evaluation sets, the reference renderer and splatting
methods receive the same physical plane.

\paragraph{Timing.}
All FPS figures are measured on a single, otherwise-idle NVIDIA A100-SXM4-80GB at
the $1600\times1600$ evaluation resolution. For each checkpoint we render all
$90$ held-out test views for three passes ($270$ frames), discard the first $30$
as warm-up, time each remaining frame with \texttt{torch.cuda.synchronize} on both
sides, and report the reciprocal of the \emph{median} per-frame time; the median
and the single-GPU run guard against transient contention. Timing includes the
full per-view forward, i.e.\ opacity conditioning and the clip factor for our
operator and the deformation-network evaluation for the ClipGS reimplementation;
it excludes disk I/O and image saving.
Consequently, the FPS comparison characterizes A100 performance at the stated
resolution. Ours/MM/HC isolate operator
cost in the shared renderer, whereas ClipGS timing remains a system-level
comparison that includes its additional network.

\paragraph{Backbones and warm start.}
All four systems receive the same per-volume initialization, using Render-FM
feed-forward predictions where available~\citep{gao2025renderfm}. The warm start
is stored as a point cloud carrying positions, spherical-harmonic color, opacity,
scale, rotation, and conditioning attributes; each backbone loads the attributes
it supports.
Ours, MM, and HC use \dgs{} as the
backbone~\citep{gao20246dgs,gao20257dgs,liu2025universal}. The Cholesky-precision
diagonal of the conditioning block is initialized at $2.0$. The ClipGS reimplementation uses
vanilla 3DGS, as in the cited paper~\citep{li2025clipgs}.

\paragraph{Optimization.}
All systems train for $30{,}000$ iterations with densification enabled,
$\lambda_{\mathrm{s}}=0.2$ in the photometric loss, and otherwise the standard
3DGS optimizer settings. The Mip-Splatting filter is enabled for the three
\dgs{} variants; the ClipGS system baseline uses vanilla 3DGS without the
filter. All tables and figures use the fixed $30{,}000$-iteration checkpoint;
the test set is reserved for final evaluation. Each training frame carries its
clipping plane $(\mathbf{n},\tau)$ in the dataset metadata, and the same plane
is passed to the rasterizer during training. Each scan--method pair is trained once with
the Python, NumPy, and PyTorch random generators fixed to seed $0$; reported
image metrics aggregate all held-out views from that fixed run. The tables
therefore report deterministic, matched fixed-run comparisons; each metric
aggregates all held-out views from that run.

\section{Baseline Implementation Details}
\label{supp:baselines}

\paragraph{ClipGS reimplementation.}
We implement ClipGS~\citep{li2025clipgs} from its published description,
following the stated vanilla-3DGS backbone, trainable binary visibility, and
plane-conditioned deformation MLP. We train it end to end
under the same initialization, data, clipping planes, and iteration budget as
the other systems. We report it as a system-level comparison; the controlled
operator conclusions come from Ours, MM, and HC on the shared N-DGS backbone.

\paragraph{RaRa integration.}
We evaluate RaRa~\citep{li2025rara} through the authors' released CUDA
rasterization kernel, driven by our cut-eval cameras and the shared fixed
checkpoints. Their plane convention keeps the side $\mathbf{n}^{\top}\mathbf{x}
+ d > 0$, so our plane $(\mathbf{n},\tau)$ is passed as $(-\mathbf{n},\tau)$, and
RaRa is rendered with the same Mip-Splatting-filtered forward as the other
operators, matching antialiasing across the comparison. We applied three
compatibility fixes to the released kernel before evaluation:
the per-primitive decay-weight buffer is initialized to one, so that primitives
classified as fully visible render unchanged rather than vanishing, and chords
whose endpoints are both invisible contribute zero; we additionally removed a
duplicated function specifier for toolchain compatibility. These fixes preserve
the clipping formula. Rendering the identical unclipped checkpoint through both
rasterizers at the cut-eval cameras gives $42.8$~dB mean PSNR, confirming
matched off-plane behavior.

\section{Cut-Eval Protocol and Metric Definitions}
\label{supp:metrics}
For the voxel-axis evaluation, we fix three clipping planes per volume, one per
voxel axis at the volume core, and render two view families of five cameras per
plane, giving $30$ views per volume, all held out from training. Perpendicular
views look down the plane normal so the exposed cut face fills the frame;
grazing views look nearly along the plane, tilted by $1.5^{\circ}$ to expose
its cut edge. The clipped reference is produced by the reference volume
renderer with its native half-space clip; the unclipped companion uses the same
camera with clipping disabled.

Table~\ref{tab:cutface-full} expands the voxel-axis average in the main paper
into per-volume values. XClipGS has the lowest Leak on every volume and the
highest \bandS{} on seven of eight. Its Leak spans $0.02$ on
\texttt{lower-limb} to $1.45$ on \texttt{knee-joint} (all $\times10^{-2}$).
This cross-volume range reflects the scene-dependent near-plane foreground
energy in the metric denominator; within \texttt{knee-joint}, XClipGS remains
below MM ($14.65$), HC ($22.44$), and ClipGS ($28.64$).

\begin{table*}[t]
\centering
\caption{\textbf{Per-volume cut-face evaluation.}
Errors and Leak are $\times10^{-2}$; \textbf{bold} marks the full-precision best.}
\label{tab:cutface-full}
\setlength{\tabcolsep}{3pt}
\resizebox{\textwidth}{!}{%
\begin{tabular}{l rccc @{\hskip 6pt} rccc @{\hskip 6pt} rccc @{\hskip 6pt} rccc}
\toprule
& \multicolumn{4}{c}{\textbf{XClipGS (Ours)}} & \multicolumn{4}{c}{ClipGS} & \multicolumn{4}{c}{MM} & \multicolumn{4}{c}{HC} \\
\cmidrule(lr){2-5}\cmidrule(lr){6-9}\cmidrule(lr){10-13}\cmidrule(lr){14-17}
Scene & \bandS\,$\uparrow$ & \cdemG\,$\downarrow$ & \cdemP\,$\downarrow$ & \leakm\,$\downarrow$ & \bandS\,$\uparrow$ & \cdemG\,$\downarrow$ & \cdemP\,$\downarrow$ & \leakm\,$\downarrow$ & \bandS\,$\uparrow$ & \cdemG\,$\downarrow$ & \cdemP\,$\downarrow$ & \leakm\,$\downarrow$ & \bandS\,$\uparrow$ & \cdemG\,$\downarrow$ & \cdemP\,$\downarrow$ & \leakm\,$\downarrow$ \\
\midrule
\texttt{abdomen}& \textbf{.899} & \textbf{28.2} & \textbf{34.4} & \textbf{0.26} & .890 & 43.1 & 39.8 & 12.93 & .895 & 31.3 & 38.6 & 3.32 & .895 & 38.9 & 34.7 & 10.50 \\
\texttt{intestine}& .920 & 22.4 & \textbf{17.2} & \textbf{0.16} & .901 & 30.9 & 17.7 & 12.68 & \textbf{.924} & \textbf{20.4} & 17.5 & 2.79 & .902 & 25.6 & 17.7 & 8.28 \\
\texttt{knee-joint}& \textbf{.825} & \textbf{15.2} & 11.1 & \textbf{1.45} & .707 & 24.7 & 12.2 & 28.64 & .793 & 15.7 & \textbf{10.7} & 14.65 & .734 & 19.6 & 11.6 & 22.44 \\
\texttt{lower-limb}& \textbf{.875} & \textbf{19.2} & 19.0 & \textbf{0.02} & .821 & 30.5 & 19.9 & 9.81 & .869 & 19.4 & \textbf{18.8} & 2.09 & .847 & 26.0 & 19.6 & 8.08 \\
\texttt{vascular}& \textbf{.876} & 25.4 & \textbf{13.1} & \textbf{0.06} & .838 & 29.1 & 14.1 & 5.19 & .868 & \textbf{24.9} & 14.2 & 0.77 & .847 & 27.9 & 13.6 & 3.79 \\
\texttt{heart}& \textbf{.877} & \textbf{21.5} & 12.2 & \textbf{0.41} & .824 & 32.5 & 12.9 & 30.68 & .871 & 24.5 & \textbf{11.8} & 16.33 & .826 & 29.2 & 12.4 & 23.00 \\
\texttt{nose}\,(MRI)& \textbf{.720} & \textbf{25.3} & 15.2 & \textbf{0.09} & .646 & 39.6 & 16.0 & 17.29 & .718 & 28.3 & \textbf{15.0} & 3.88 & .665 & 34.0 & 15.6 & 9.62 \\
\texttt{hand}\,(MRI)& \textbf{.886} & 24.6 & 20.9 & \textbf{0.71} & .848 & 36.3 & 23.2 & 14.12 & .883 & \textbf{24.2} & \textbf{20.7} & 5.60 & .838 & 34.1 & 22.7 & 14.63 \\
\midrule
\texttt{average}& \textbf{.860} & \textbf{22.7} & \textbf{17.9} & \textbf{0.40} & .809 & 33.4 & 19.5 & 16.42 & .853 & 23.6 & 18.4 & 6.18 & .819 & 29.4 & 18.5 & 12.54 \\
\bottomrule
\end{tabular}}
\end{table*}

\paragraph{Arbitrary-normal generalization.}
The main benchmark varies plane offsets along the three voxel axes. To test
orientation generalization separately, we sample five deterministic
physical-world normals per volume, rejecting any normal within $20^\circ$ of a
voxel axis or within $20^\circ$ of another sampled orientation. Sampling is
performed once with seed $27$. Every plane
passes through the volume center and has one perpendicular and two grazing
views, for $15$ views per volume. We render matched clipped and unclipped
references and evaluate the original checkpoints without retraining.
Table~\ref{tab:cutface-oblique-full} gives the complete results. XClipGS has the
lowest Leak on all eight volumes, the lowest \cdemG{} on seven, and the highest
\bandS{} on six. MM has the lowest \cdemG{} on \texttt{vascular} and the lowest
\cdemP{} on four volumes, but XClipGS remains best on the average of every
reported metric.

\begin{table*}[t]
\centering
\caption{\textbf{Per-volume arbitrary-normal evaluation.}
Errors and Leak are $\times10^{-2}$; \textbf{bold} marks the full-precision best.}
\label{tab:cutface-oblique-full}
\setlength{\tabcolsep}{3pt}
\resizebox{\textwidth}{!}{%
\begin{tabular}{l rccc @{\hskip 6pt} rccc @{\hskip 6pt} rccc @{\hskip 6pt} rccc}
\toprule
& \multicolumn{4}{c}{\textbf{XClipGS (Ours)}} & \multicolumn{4}{c}{ClipGS} & \multicolumn{4}{c}{MM} & \multicolumn{4}{c}{HC} \\
\cmidrule(lr){2-5}\cmidrule(lr){6-9}\cmidrule(lr){10-13}\cmidrule(lr){14-17}
Scene & \bandS\,$\uparrow$ & \cdemG\,$\downarrow$ & \cdemP\,$\downarrow$ & \leakm\,$\downarrow$ & \bandS\,$\uparrow$ & \cdemG\,$\downarrow$ & \cdemP\,$\downarrow$ & \leakm\,$\downarrow$ & \bandS\,$\uparrow$ & \cdemG\,$\downarrow$ & \cdemP\,$\downarrow$ & \leakm\,$\downarrow$ & \bandS\,$\uparrow$ & \cdemG\,$\downarrow$ & \cdemP\,$\downarrow$ & \leakm\,$\downarrow$ \\
\midrule
\texttt{abdomen}& \textbf{.924} & \textbf{24.9} & \textbf{26.3} & \textbf{0.27} & .912 & 39.3 & 31.7 & 15.62 & .922 & 27.8 & 28.7 & 3.80 & .901 & 39.0 & 27.4 & 17.52 \\
\texttt{intestine}& .918 & \textbf{18.6} & \textbf{12.7} & \textbf{0.11} & .914 & 31.5 & 13.1 & 14.43 & \textbf{.925} & 22.3 & 12.9 & 2.94 & .904 & 31.6 & 13.1 & 16.14 \\
\texttt{knee-joint}& \textbf{.901} & \textbf{18.8} & 10.0 & \textbf{0.16} & .849 & 32.0 & 11.8 & 28.64 & .884 & 22.2 & \textbf{9.8} & 12.90 & .852 & 32.2 & 10.7 & 31.06 \\
\texttt{lower-limb}& .903 & \textbf{19.5} & 8.7 & \textbf{0.02} & .884 & 29.5 & 9.7 & 8.95 & \textbf{.907} & 20.3 & \textbf{8.6} & 2.02 & .899 & 29.3 & 8.7 & 9.67 \\
\texttt{vascular}& \textbf{.881} & 34.7 & \textbf{9.2} & \textbf{0.03} & .847 & 35.9 & 10.5 & 5.27 & .878 & \textbf{26.0} & 14.0 & 0.98 & .860 & 35.3 & 9.6 & 5.75 \\
\texttt{heart}& \textbf{.846} & \textbf{21.8} & 14.2 & \textbf{0.05} & .797 & 28.1 & 14.7 & 26.00 & .841 & 22.1 & \textbf{13.8} & 11.65 & .805 & 30.3 & 14.5 & 29.44 \\
\texttt{nose}\,(MRI)& \textbf{.806} & \textbf{24.9} & \textbf{17.2} & \textbf{0.11} & .755 & 44.5 & 17.8 & 17.88 & .790 & 30.3 & 17.3 & 4.45 & .747 & 44.2 & 18.1 & 17.61 \\
\texttt{hand}\,(MRI)& \textbf{.815} & \textbf{24.2} & 16.4 & \textbf{0.26} & .790 & 46.5 & 16.6 & 21.12 & .812 & 32.7 & \textbf{15.9} & 6.58 & .769 & 45.7 & 16.8 & 19.18 \\
\midrule
\texttt{average}& \textbf{.874} & \textbf{23.5} & \textbf{14.3} & \textbf{0.13} & .844 & 35.9 & 15.7 & 17.24 & .870 & 25.5 & 15.1 & 5.67 & .842 & 35.9 & 14.9 & 18.30 \\
\bottomrule
\end{tabular}}
\end{table*}

For the near-edge coordinate in a grazing view, we use the deterministic
least-squares line proxy
$\mathbf{l}=\mathrm{pinv}(\mathbf{P})^{\top}[\mathbf{n};-\tau]$, where
$\mathbf{P}$ is the camera projection matrix. Equivalently, $\mathbf{l}$
minimizes $\|\mathbf{P}^{\top}\mathbf{l}-[\mathbf{n};-\tau]\|_2$. A general
3D plane projects nonlinearly under perspective, so this proxy supplies a stable
near-edge coordinate for our cameras placed $1.5^{\circ}$ from edge-on. Its sign is oriented so
reference foreground adjacent to the cut lies at $s<0$, independently of the
stored plane orientation. Grazing metrics use a
$\pm12$-pixel strip around this line; perpendicular metrics use the exposed-face
foreground.

\paragraph{Notation.}
All image metrics operate on scalar luminance
$Y(\mathbf{x})=\tfrac13\sum_{c}I_c(\mathbf{x})\in[0,1]$ (mean of the three
$[0,1]$ channels) at pixel $\mathbf{x}$. The reference renderer uses a black
background, so the foreground mask is
$\mathcal{F}(I)=\{\mathbf{x}: \max_c I_c(\mathbf{x})>\theta_{\mathrm{fg}}\}$ with
$\theta_{\mathrm{fg}}=0.04$. Write $s(\mathbf{x})$ for the signed pixel distance
to the near-edge line proxy, oriented so $s<0$ is the kept side. Two band
restrictions appear below: the \emph{perpendicular band} is the reference
foreground $\mathcal{F}(G)$ for band-fidelity metrics, and the whole frame for the
region-count \cdem{} (which is already gated by the affected regions $A_G,A_R$);
the \emph{grazing band} is the $\pm12$-pixel strip $\{|s|\le12\}$ about the line.
The definitions below apply unchanged to either evaluation set: three
voxel-axis planes or five arbitrary-normal planes.

\paragraph{Band fidelity.}
\bandS{} is the SSIM between the method render and the reference over the
reference foreground $\mathcal{F}(G)$, computed per perpendicular view and
averaged over the perpendicular views of all planes in the evaluated set. We
also compute band PSNR but omit it
from the main tables, as the operators differ by hundredths of a decibel and
leakage can offset a small brightness bias.

\paragraph{Difference-referenced cut error (\cdem).}
For a camera we form the \emph{removal maps} of the method render
$R_{\mathrm{full}},R_{\mathrm{clip}}$ and the reference
$G_{\mathrm{full}},G_{\mathrm{clip}}$ (same camera, clip on and off):
\begin{equation}
D_R = Y(R_{\mathrm{full}})-Y(R_{\mathrm{clip}}),\quad
D_G = Y(G_{\mathrm{full}})-Y(G_{\mathrm{clip}}).
\end{equation}
Thresholding at $\theta=0.04$ gives the binary affected regions
$A_R=\{|D_R|>\theta\}$ and $A_G=\{|D_G|>\theta\}$, and
\begin{equation}
\mcdem = \frac{\sum_v|(A_G\setminus A_R)\cap\mathcal{B}| + \sum_v|(A_R\setminus A_G)\cap\mathcal{B}|}
             {\sum_v|A_G\cap\mathcal{B}|},
\end{equation}
the region symmetric difference (under-removal plus over-removal) normalized by
the reference region, with the pixel counts summed over all views $v$ of the
family and all planes in the evaluated set before a single ratio is taken. Differencing
each render against its own unclipped companion cancels the reconstruction floor
that a direct clipped-vs-reference comparison carries, and the set difference
prevents under-removal (a hole the method leaves unfilled) from being paid for by
over-removal (material it wrongly deletes). For \cdemP{} the band $\mathcal{B}$ is
the whole perpendicular frame (the $A_G,A_R$ gating already localizes the score to
the affected face); for \cdemG{} it is the grazing $\pm12$-pixel strip.

\paragraph{Leak.}
On grazing views, \leakm{} is the fraction of the method's near-plane foreground
energy that falls on the culled side where the reference is background:
\begin{equation}
\mleak = \frac{\sum_{\mathbf{x}\in \mathcal{L}} \max_c R_c(\mathbf{x})}
              {\sum_{\mathbf{x}\in \mathcal{N}} \max_c R_c(\mathbf{x})},
\end{equation}
where $\mathcal{N}=\mathcal{F}(R)\cap\{|s|\le 60\}$ is the method's near-plane
foreground and
$\mathcal{L}=\mathcal{N}\cap\{s>2\}\setminus\mathcal{F}(G)$
restricts it to just past the cut edge, on the culled side, and off the reference
foreground (where the reference is empty). The numerator and denominator are
summed over all grazing views of the evaluated set before the ratio is taken
(pooling avoids the instability of averaging per-view ratios on thin grazing
bands). Exact half-space truncation introduces no wrong-side mass in the
represented Gaussian model. Image-space \leakm{} uses an independently rendered
reference mask and therefore measures proxy reconstruction and renderer mismatch
alongside the soft tails or overshooting culls introduced by approximate
operators.
We interpret \leakm{} jointly with \bandS{} and \cdem{}: together they couple
culled-side energy with retained-face fidelity and over-removal.

\paragraph{Camera-independent geometric cut error (\cerrD).}
This diagnostic is computed in closed form from the primitives. For plane
$(\mathbf{n},\tau)$ and primitive $(\boldsymbol{\mu},\boldsymbol{\Sigma},\alpha)$,
let $t=(\tau-\mathbf{n}^{\top}\boldsymbol{\mu})/\sqrt{\mathbf{n}^{\top}\boldsymbol{\Sigma}\mathbf{n}}$;
the exact kept opacity mass is $e_i=\alpha_i\,\Phi(t_i)$. We define each
operator's misplaced mass directly. Ours has $\epsilon_i^{\mathrm{Ours}}=0$ by
the analytic definition, making \cerrD{} a camera-independent operator-geometry
diagnostic; Section~\ref{supp:impl} independently validates the finite-precision
CUDA path.
For HC,
\begin{equation}
\epsilon_i^{\mathrm{HC}}
=\left|\alpha_i\mathds{1}[t_i\geq0]-e_i\right|.
\end{equation}
This is removed kept mass when HC drops the primitive and retained culled mass
when it keeps it. Because MM's total opacity mass is $e_i$, its misplaced mass
is the moment-matched Gaussian tail beyond the plane,
\begin{equation}
\epsilon_i^{\mathrm{MM}}
=e_i\,\Phi\!\left(-\frac{t_i+\lambda(t_i)}{\sqrt{v(t_i)}}\right).
\end{equation}
For the voxel-axis diagnostic in Table~\ref{tab:cerr3d-full}, we pool the three
planes and normalize by the exact kept mass:
\begin{equation}
\mcerrD = \frac{\sum_i \epsilon_i}{\sum_i e_i}.
\end{equation}
$\boldsymbol{\Sigma}$ is the full rotated covariance with the persisted
Mip-Splatting filter, and $t$ uses the exact evaluation-plane normals and offsets.

Table~\ref{tab:cerr3d-full} reports the diagnostic on the Ours-trained and
HC-trained interiors. MM can place only a soft tail beyond the plane, whereas HC
can both remove kept mass and retain culled mass. This controlled diagnostic
compares render-time rules on the shared N-DGS interiors; ClipGS remains in the
image-space system comparison.
On the Ours-trained interior, MM and HC average $0.225\times10^{-2}$ and
$1.400\times10^{-2}$;
on the HC-trained interior, they average $0.213\times10^{-2}$ and
$1.317\times10^{-2}$. The ordering is therefore unchanged by the checkpoint.

\begin{table}[t]
\centering
\caption{\textbf{Camera-independent cut error.}
CErr\textsubscript{3D}$\downarrow$ ($\times10^{-2}$) on both interiors;
Ours is \textbf{0.000} by construction.}
\label{tab:cerr3d-full}
\setlength{\tabcolsep}{4pt}
\resizebox{\columnwidth}{!}{%
\begin{tabular}{lc@{\hskip 8pt}cc@{\hskip 8pt}cc}
\toprule
& \multicolumn{1}{c}{Both}
& \multicolumn{2}{c}{Ours-trained}
& \multicolumn{2}{c}{HC-trained} \\
\cmidrule(lr){2-2}\cmidrule(lr){3-4}\cmidrule(lr){5-6}
Scene & Ours & MM & HC & MM & HC \\
\midrule
\texttt{abdomen}    & \textbf{0.000} & 0.166 & 1.015 & 0.166 & 1.015 \\
\texttt{intestine}  & \textbf{0.000} & 0.191 & 1.201 & 0.196 & 1.210 \\
\texttt{knee-joint} & \textbf{0.000} & 0.217 & 1.347 & 0.204 & 1.249 \\
\texttt{lower-limb} & \textbf{0.000} & 0.205 & 1.264 & 0.196 & 1.183 \\
\texttt{vascular}   & \textbf{0.000} & 0.216 & 1.373 & 0.216 & 1.367 \\
\texttt{heart}      & \textbf{0.000} & 0.149 & 0.931 & 0.140 & 0.862 \\
\texttt{nose} (MRI) & \textbf{0.000} & 0.227 & 1.393 & 0.224 & 1.379 \\
\texttt{hand} (MRI) & \textbf{0.000} & 0.426 & 2.675 & 0.361 & 2.274 \\
\midrule
\texttt{avg}        & \textbf{0.000} & 0.225 & 1.400 & 0.213 & 1.317 \\
\bottomrule
\end{tabular}
}
\end{table}

\paragraph{Arbitrary-normal geometric error.}
On the five arbitrary-normal planes per volume, XClipGS retains zero
CErr\textsubscript{3D} by construction, whereas MM and HC average
$0.174\times10^{-2}$ and $1.083\times10^{-2}$, respectively. Thus the
geometric ordering is not specific to voxel-axis planes.

\paragraph{Metric-parameter sensitivity.}
For the voxel-axis set, we recompute both image-space metrics from fixed
renders over a parameter grid. For \cdemG{}, the affected-region threshold is
$\theta\in\{0.02,0.04,0.06,0.08\}$ and the grazing-band half-width is
$b\in\{8,12,16,24\}$ pixels, giving $16$ combinations; \cdemP{} uses the four
thresholds and the full face. For \leakm{}, we take the Cartesian product of
foreground threshold $\theta_{\mathrm{fg}}\in\{0.02,0.04,0.08\}$, cut-edge
margin $\delta\in\{1,2,4\}$ pixels, and near-plane half-width
$w\in\{40,60,80\}$ pixels, giving $27$ combinations. Each setting uses the
same pooled-within-scene, mean-across-scenes aggregation as the main tables.

Table~\ref{tab:metric-sensitivity} reports eight-volume ranges and per-method
win counts. Ours has the lowest face CDE at all
four thresholds and the lowest Leak in all $27$ settings. Edge CDE is the close
comparison: Ours is best in $11$ of $16$ settings and MM in the other five;
ClipGS and HC are never best. Thus the face-localization and leakage conclusions
are stable, while the small default edge-CDE advantage over MM is
parameter-dependent. The main-table row uses
$(\theta,b,\theta_{\mathrm{fg}},\delta,w)=(0.04,12,0.04,2,60)$.

\begin{table*}[t]
\centering
\caption{\textbf{Metric-parameter sensitivity.} Ranges of eight-volume means
($\times10^{-2}$); ``Wins'' counts settings with the lowest mean.}
\label{tab:metric-sensitivity}
\setlength{\tabcolsep}{5pt}
\begin{tabular}{l c cccc l}
\toprule
Metric & Settings & Ours & ClipGS & MM & HC & Wins \\
\midrule
\cdemG{} & 16 & 17.47--29.57 & 21.73--50.55 & 15.58--36.06 & 19.24--43.64
  & Ours 11; MM 5 \\
\cdemP{} & 4 & 11.97--25.74 & 13.18--28.33 & 12.14--26.36 & 12.42--27.01
  & Ours 4 \\
\leakm{} & 27 & 0.20--0.72 & 12.60--20.96 & 4.24--8.64 & 9.56--16.05
  & Ours 27 \\
\bottomrule
\end{tabular}
\end{table*}

\section{External-View-Only Ablation}
\label{supp:extonly}
Table~\ref{tab:extonly-full} gives the per-scan breakdown of the
clip-aware-supervision ablation. The clip-aware model is our exact operator
trained on the full view set; the external-view-only model is the identical
configuration trained on the intact views alone, with every clipped training
view removed. Both are scored on the same held-out split at the fixed
$30{,}000$-iteration checkpoint (so CA reproduces the main-paper ``Ours''
column), reported here as
intact-, clipped-, and all-view PSNR alongside the cut-face metrics for both
models. Clip-aware supervision raises clipped-view PSNR on all eight scans,
improving the average by $8.26$~dB ($23.76\!\to\!32.03$), and substantially
improves band SSIM and \cdemP{}. EX averages $0.72$~dB higher on intact views,
while the full model's gain concentrates on the newly exposed cross-section.
Leak stays comparably low because both models share the exact operator; band
SSIM and CDE capture the interior-fidelity gain from clipped supervision.
The cut-face columns use the voxel-axis evaluation set.

\begin{table*}[t]
\centering
\caption{\textbf{Clip-aware supervision ablation.} CA uses full supervision;
EX uses intact views only. iP, cP, and aP are PSNR on intact, clipped, and all views;
cut errors and Leak are $\times10^{-2}$.}
\label{tab:extonly-full}
\setlength{\tabcolsep}{3pt}
\resizebox{\textwidth}{!}{%
\begin{tabular}{l ccccccc @{\hskip 10pt} ccccccc}
\toprule
& \multicolumn{7}{c}{\textbf{CA} = Ours (full supervision)}
& \multicolumn{7}{c}{EX = ablation (external views only)} \\
\cmidrule(lr){2-8}\cmidrule(lr){9-15}
Scene & iP\,$\uparrow$ & cP\,$\uparrow$ & aP\,$\uparrow$ & \bandS\,$\uparrow$ & \leakm\,$\downarrow$ & \cdemG\,$\downarrow$ & \cdemP\,$\downarrow$
      & iP\,$\uparrow$ & cP\,$\uparrow$ & aP\,$\uparrow$ & \bandS\,$\uparrow$ & \leakm\,$\downarrow$ & \cdemG\,$\downarrow$ & \cdemP\,$\downarrow$ \\
\midrule
\texttt{abdomen} & 33.55 & 33.05 & 33.30 & 0.899 & 0.26 & 28.2 & 34.4 & 35.03 & 23.01 & 29.02 & 0.325 & 0.79 & 39.8 & 51.8 \\
\texttt{intestine} & 32.59 & 30.96 & 31.77 & 0.920 & 0.16 & 22.4 & 17.2 & 33.58 & 20.84 & 27.21 & 0.340 & 0.27 & 22.8 & 24.3 \\
\texttt{knee-joint} & 33.52 & 24.76 & 29.14 & 0.825 & 1.45 & 15.2 & 11.1 & 33.62 & 18.40 & 26.01 & 0.244 & 2.16 & 18.3 & 17.3 \\
\texttt{lower-limb} & 34.53 & 34.04 & 34.28 & 0.875 & 0.02 & 19.2 & 19.0 & 35.35 & 26.56 & 30.95 & 0.555 & 0.03 & 21.1 & 24.4 \\
\texttt{vascular} & 34.30 & 38.46 & 36.38 & 0.876 & 0.06 & 25.4 & 13.1 & 34.75 & 31.19 & 32.97 & 0.416 & 0.08 & 27.5 & 20.4 \\
\texttt{heart} & 31.04 & 27.37 & 29.20 & 0.877 & 0.41 & 21.5 & 12.2 & 31.47 & 18.58 & 25.03 & 0.263 & 0.39 & 22.8 & 16.8 \\
\texttt{nose} (MRI) & 38.21 & 33.63 & 35.92 & 0.720 & 0.09 & 25.3 & 15.2 & 39.03 & 25.74 & 32.39 & 0.198 & 0.10 & 28.4 & 20.7 \\
\texttt{hand} (MRI) & 43.02 & 33.95 & 38.48 & 0.886 & 0.71 & 24.6 & 20.9 & 43.72 & 25.77 & 34.74 & 0.425 & 0.57 & 25.3 & 28.8 \\
\midrule
\texttt{avg} & 35.09 & 32.03 & 33.56 & 0.860 & 0.40 & 22.7 & 17.9 & 35.82 & 23.76 & 29.79 & 0.346 & 0.55 & 25.8 & 25.6 \\
\bottomrule
\end{tabular}}
\end{table*}

\section{Operator-Swap Controls and Per-Volume Results}
\label{supp:fairness}
All operators use the same voxel-axis planes and cameras as the main
operator-swap comparison.

\paragraph{Control design.}
We assess operator/checkpoint coupling in three configurations. First, the
trained-system comparison scores HC and MM on their own co-adapted interiors.
Second, the fixed-interior comparison supplies every render-time rule, including
RaRa, with the same clip-aware interior. Third,
Table~\ref{tab:opswap-hc-full} repeats the swap on the HC-trained checkpoint.
Ours retains its average edge-error and leakage lead across the co-adapted and
two fixed-interior comparisons.

\paragraph{Per-volume results.}
\label{supp:pervol}
The main paper reports average operator-swap results on two fixed interiors.
Tables~\ref{tab:opswap-full} and~\ref{tab:opswap-hc-full} provide the corresponding
per-volume measurements. In each table, every operator renders the same
primitives with matched cameras, clipping planes, and Mip-Splatting forward.
Metrics and units follow the main paper: \bandS\,$\uparrow$ is cut-face band
SSIM, \cdemG/\cdemP\,$\downarrow$ are difference-referenced cut errors at the
edge and over the exposed face, and \leakm\,$\downarrow$ is culled-side leaked
energy. Error and leakage values are $\times10^{-2}$.

On the Ours-trained interior, HC has the largest edge error \cdemG{} on all
eight volumes, and Ours leaks least on all eight; on the HC-trained interior the
same ordering holds, establishing consistency across both fixed interiors.

\begin{table*}[t]
\centering
\caption{\textbf{Operator swap on the Ours-trained interior.}
\textbf{Bold} marks the full-precision best.}
\label{tab:opswap-full}
\setlength{\tabcolsep}{3pt}
\resizebox{\textwidth}{!}{%
\begin{tabular}{l rccc @{\hskip 6pt} rccc @{\hskip 6pt} rccc @{\hskip 6pt} rccc}
\toprule
& \multicolumn{4}{c}{\textbf{Ours (analytic)}} & \multicolumn{4}{c}{RaRa} & \multicolumn{4}{c}{MM} & \multicolumn{4}{c}{HC} \\
\cmidrule(lr){2-5}\cmidrule(lr){6-9}\cmidrule(lr){10-13}\cmidrule(lr){14-17}
Scene & \bandS\,$\uparrow$ & \cdemG\,$\downarrow$ & \cdemP\,$\downarrow$ & \leakm\,$\downarrow$ & \bandS\,$\uparrow$ & \cdemG\,$\downarrow$ & \cdemP\,$\downarrow$ & \leakm\,$\downarrow$ & \bandS\,$\uparrow$ & \cdemG\,$\downarrow$ & \cdemP\,$\downarrow$ & \leakm\,$\downarrow$ & \bandS\,$\uparrow$ & \cdemG\,$\downarrow$ & \cdemP\,$\downarrow$ & \leakm\,$\downarrow$ \\
\midrule
\texttt{abdomen}& .899 & 28.2 & 34.4 & \textbf{0.26} & \textbf{.901} & \textbf{28.0} & \textbf{34.3} & 1.31 & .895 & 31.0 & 35.1 & 3.05 & .892 & 39.1 & 34.6 & 11.06 \\
\texttt{intestine}& \textbf{.920} & 22.4 & \textbf{17.2} & \textbf{0.16} & \textbf{.920} & 22.6 & 17.3 & 5.65 & .917 & \textbf{20.9} & \textbf{17.2} & 2.65 & .907 & 26.1 & 17.5 & 9.37 \\
\texttt{knee-joint}& \textbf{.825} & 15.2 & 11.1 & \textbf{1.45} & .823 & \textbf{15.0} & 11.1 & 6.00 & .805 & 15.9 & \textbf{11.0} & 15.31 & .798 & 18.5 & 11.4 & 26.88 \\
\texttt{lower-limb}& \textbf{.875} & \textbf{19.2} & \textbf{19.0} & \textbf{0.02} & .872 & \textbf{19.2} & 19.6 & 7.95 & .864 & 19.8 & \textbf{19.0} & 2.48 & .862 & 27.8 & 19.3 & 10.69 \\
\texttt{vascular}& \textbf{.876} & 25.4 & \textbf{13.1} & \textbf{0.06} & .857 & 27.8 & 14.3 & 6.68 & .867 & \textbf{25.1} & \textbf{13.1} & 0.74 & .864 & 28.5 & 13.2 & 3.70 \\
\texttt{heart}& .877 & \textbf{21.5} & \textbf{12.2} & \textbf{0.41} & \textbf{.878} & 21.9 & 12.5 & 6.62 & .859 & 23.6 & 12.4 & 13.39 & .846 & 27.6 & 12.5 & 25.93 \\
\texttt{nose}\,(MRI)& \textbf{.720} & \textbf{25.3} & \textbf{15.2} & \textbf{0.09} & .715 & 27.1 & 15.5 & 6.80 & .698 & 28.2 & 15.3 & 4.41 & .678 & 34.1 & 15.4 & 12.60 \\
\texttt{hand}\,(MRI)& \textbf{.886} & \textbf{24.6} & \textbf{20.9} & \textbf{0.71} & .882 & 25.8 & 21.4 & 8.94 & .877 & 25.1 & 21.2 & 4.75 & .854 & 39.0 & 21.6 & 18.91 \\
\midrule
\texttt{avg}& \textbf{.860} & \textbf{22.7} & \textbf{17.9} & \textbf{0.40} & .856 & 23.4 & 18.3 & 6.24 & .848 & 23.7 & 18.0 & 5.85 & .838 & 30.1 & 18.2 & 14.89 \\
\bottomrule
\end{tabular}}
\end{table*}

\begin{table*}[t]
\centering
\caption{\textbf{Operator swap on the HC-trained interior.}
\textbf{Bold} marks the full-precision best.}
\label{tab:opswap-hc-full}
\setlength{\tabcolsep}{3pt}
\resizebox{\textwidth}{!}{%
\begin{tabular}{l rccc @{\hskip 6pt} rccc @{\hskip 6pt} rccc @{\hskip 6pt} rccc}
\toprule
& \multicolumn{4}{c}{\textbf{Ours (analytic)}} & \multicolumn{4}{c}{RaRa} & \multicolumn{4}{c}{MM} & \multicolumn{4}{c}{HC} \\
\cmidrule(lr){2-5}\cmidrule(lr){6-9}\cmidrule(lr){10-13}\cmidrule(lr){14-17}
Scene & \bandS\,$\uparrow$ & \cdemG\,$\downarrow$ & \cdemP\,$\downarrow$ & \leakm\,$\downarrow$ & \bandS\,$\uparrow$ & \cdemG\,$\downarrow$ & \cdemP\,$\downarrow$ & \leakm\,$\downarrow$ & \bandS\,$\uparrow$ & \cdemG\,$\downarrow$ & \cdemP\,$\downarrow$ & \leakm\,$\downarrow$ & \bandS\,$\uparrow$ & \cdemG\,$\downarrow$ & \cdemP\,$\downarrow$ & \leakm\,$\downarrow$ \\
\midrule
\texttt{abdomen}& .895 & 28.8 & \textbf{34.2} & \textbf{0.29} & .896 & \textbf{28.7} & \textbf{34.2} & 1.28 & \textbf{.897} & 31.3 & 34.8 & 3.39 & .895 & 38.9 & 34.7 & 10.50 \\
\texttt{intestine}& .909 & 22.3 & 17.6 & \textbf{0.16} & .909 & 22.6 & 17.7 & 4.82 & \textbf{.910} & \textbf{20.9} & \textbf{17.4} & 3.20 & .902 & 25.6 & 17.7 & 8.28 \\
\texttt{knee-joint}& .761 & 15.9 & 11.9 & \textbf{1.36} & \textbf{.772} & 15.0 & 12.0 & 4.89 & .755 & \textbf{14.9} & \textbf{11.6} & 12.69 & .734 & 19.6 & \textbf{11.6} & 22.44 \\
\texttt{lower-limb}& \textbf{.850} & 19.4 & 19.7 & \textbf{0.02} & .845 & 19.3 & 20.3 & 6.05 & .846 & \textbf{19.1} & \textbf{19.4} & 1.75 & .847 & 26.0 & 19.6 & 8.08 \\
\texttt{vascular}& \textbf{.848} & 25.6 & 13.7 & \textbf{0.06} & .829 & 27.6 & 14.8 & 5.15 & .847 & \textbf{25.2} & \textbf{13.6} & 0.72 & .847 & 27.9 & \textbf{13.6} & 3.79 \\
\texttt{heart}& .839 & \textbf{22.5} & 12.5 & \textbf{0.36} & .839 & 22.8 & 12.9 & 5.45 & \textbf{.844} & 25.2 & \textbf{12.4} & 12.37 & .826 & 29.2 & \textbf{12.4} & 23.00 \\
\texttt{nose}\,(MRI)& .672 & \textbf{25.5} & 15.7 & \textbf{0.09} & .667 & 26.9 & 15.9 & 4.87 & \textbf{.673} & 27.8 & \textbf{15.5} & 3.76 & .665 & 34.0 & 15.6 & 9.62 \\
\texttt{hand}\,(MRI)& .851 & 24.7 & 22.7 & \textbf{0.66} & .852 & 26.0 & 23.0 & 6.61 & \textbf{.853} & \textbf{23.4} & \textbf{22.5} & 3.09 & .838 & 34.1 & 22.7 & 14.63 \\
\midrule
\texttt{avg}& \textbf{.828} & \textbf{23.1} & 18.5 & \textbf{0.37} & .826 & 23.6 & 18.8 & 4.89 & \textbf{.828} & 23.5 & \textbf{18.4} & 5.12 & .819 & 29.4 & 18.5 & 12.54 \\
\bottomrule
\end{tabular}}
\end{table*}

\section{Supplementary Demonstrations}
\label{supp:demos}
Two demonstrations accompany this paper on the project
page.\footnote{\url{https://gaozhongpai.github.io/XClipGS/}} Neither is
required to assess the paper: every reported number comes from the CUDA implementation and
protocol described in Sections~\ref{supp:impl} and~\ref{supp:repro}, and the
main paper is self-contained. Both show behavior that is inherently temporal or
interactive and therefore cannot be conveyed by a static figure.

\paragraph{Video demonstration.}
Two screen recordings of the same held-out volume (\texttt{intestine}, CT)
render interactively while a clipping plane sweeps through the anatomy, one
using our analytic operator and one using the ClipGS reimplementation. Both are
captured live from the trained checkpoints and are unedited. They show the
temporal behavior that the still figures cannot: with the analytic operator the
exposed face stays sharp and no material appears on the culled side, whereas the
baseline leaks past the plane and its boundary flickers as whole primitives are
kept or dropped. The measured culled-side leakage on this volume differs by
roughly two orders of magnitude ($0.16$ versus $12.68\times10^{-2}$;
Table~\ref{tab:cutface-full}). Frame rates visible in the recordings come from a
shared GPU and are not the timing measurements of Section~\ref{supp:repro}.

\paragraph{Interactive demonstration.}
A self-contained web page renders a Gaussian scene and switches among the three
clip operators at render time, with a movable plane and a sweep animation, so
that hard-cull popping, the moment-matched tail, and the analytic boundary can
be compared directly. It runs offline with no build step. This page is an
independent WebGL reimplementation of the three operators for illustration; it
is \emph{not} the CUDA rasterizer used for any reported result, and it
corroborates only the qualitative operator behavior described in the
half-space clipping operators section of the main paper. The third-party rendering
engine is bundled, so the page needs no network access.

\end{document}